\documentclass[preprint,review,12pt]{elsarticle}

\usepackage{amssymb}
\usepackage{amsmath}
\usepackage{dsfont}
\usepackage{xurl}
\usepackage{float}
\usepackage{arydshln}
\usepackage{xcolor}
\usepackage{subfigure}
\usepackage{makeidx}
\makeindex

\journal{Pattern Recognition}
\usepackage{lineno}
\begin{document}
\printindex
\begin{frontmatter}



\title{Progressive Masked Refinement Self-supervised Learning for Low-Dose CT Denoising}


\author[label1]{Yichao Liu} 
\author[label5]{Zongru Shao}
\author[label6]{Rui Wen}
\author[label2,label3]{Yueyang Teng}
\author[label4]{Junwen Guo\corref{cor1}}
\ead{junwen.guo@umu.se}
\cortext[cor1]{Corresponding author}

\affiliation[label1]{organization={IWR, Heidelberg University},
	            city={Heidelberg},
	            postcode={69120}, 
	            state={Baden Württemberg},
	            country={Germany}}

\affiliation[label5]{organization={Silicon Austria Labs},
		city={Linz},
		postcode={4040}, 
	    state={Upper Austria},
		country={Austria}} 

\affiliation[label6]{organization={Institute of Science Tokyo},
	addressline={}, 
		city={Tokyo}, 
		country={Japan}} 
        
\affiliation[label2]{organization={College of Medicine and Biological Information Engineering, Northeastern University},
	            city={Shenyang},
	            postcode={110169}, 
	            state={Liaoning},
	            country={China}}
\affiliation[label3]{organization={Key Laboratory of Intelligent Computing in Medical Image, Ministry of Education},
	            city={Shenyang},
	            postcode={110169}, 
	            state={Liaoning},
	            country={China}}     

\affiliation[label4]{organization={Department of Epidemiology \& Global Health, Umeå University},
	addressline={}, 
		city={Umeå},
		postcode={90187}, 
		country={Sweden}}

\begin{abstract}
Self-supervised learning has been increasingly investigated for low-dose computed tomography (LDCT) image denoising, as it alleviates the dependence on paired normal-dose CT (NDCT) data, which are often difficult to obtain. However, many existing self-supervised blind-spot denoising methods may under-utilize pixel-wise supervisory information due to evaluating the training loss only at masked locations. To address this, we propose a Progressive Masked Refinement Learning framework that recovers NDCT features through iterative refinement with different masking patterns: the denoised output of each step is fed back as the input to the next, so that loss is calculated at a larger proportion of image pixels and the estimate is progressively driven closer to the underlying clean image. Two components make this refinement effective. First, we explicitly inject a combination of controlled Gaussian and Poisson noise during training to regularize the denoising process. Second, we introduce a step-wise mask denoising mechanism that gradually reduces the discrepancy between synthetic corruption and the noise characteristics of LDCT images. The experiments on the Mayo LDCT dataset demonstrate that the proposed method outperforms existing self-supervised approaches and achieves performance comparable to, or better than, several representative supervised denoising methods. Code is available at: \url{https://github.com/Christian-lyc/Progressive-self-supervised-blind-spot-denoising-method-for-LDCT-denoising.git}.
\end{abstract}


\begin{highlights}

\item Propose a progressive Masked Refinement Learning for low-dose CT denoising.
\item Address the underutilization of pixel-wise supervisory information loss in blind-spot denoising methods with the step-wise masking and denoising strategy.
\item Mitigate trivial identity mapping and improve pixel-correlation learning using noise-based regularization with controlled noise injection in both inputs and targets.
\item Experiments show architecture-agnostic improvements, matching or surpassing supervised low-dose CT denoising methods.
\end{highlights}


\end{frontmatter}



\section{Introduction}
\label{introduction}
Low-dose computed tomography (LDCT) has been widely adopted across diverse clinical settings, including cancer screening \cite{Chang2022RiskAOA, Chao2023ArtificialIAA}, liver disease evaluation \cite{Choi2022ProspectiveEOA, Zakharova2023HepaticSDA}, and detection of pulmonary nodules \cite{Huber2016PerformanceOUA, Kim2020LowdoseCCA}. These applications demand high image quality to ensure accurate diagnosis and treatment planning, yet the reduced radiation dose inherent to LDCT often leads to increased noise and artifacts that compromise image clarity. Effective noise reduction is therefore critical to enhance the diagnostic confidence of LDCT scans, enabling reliable visualization of subtle anatomical details and pathological changes while maintaining patient safety. 

Before deep learning, classical model-based methods such as Block-Matching 3D (BM3D) \cite{dabov2007image} suppress noise by exploiting non-local self-similarity through grouping similar patches and performing collaborative filtering in a transformed domain. The denoised images are usually blurred, although they do not require data to train the model. After that, extensive efforts have been devoted to LDCT image denoising in recent years, with supervised learning methods remaining the dominant paradigm due to their strong performance. Early representative approaches include convolutional neural network (CNN) based models such as U-Net \cite{ronneberger2015u} and Residual Encoder-Decoder CNN (RED-CNN) \cite{chen2017low}, as well as generative frameworks such as CycleGANs \cite{zhu2017unpaired, li2019low}. More recently, denoising architectures have evolved toward richer global modeling and multi-scale representation learning, exemplified by transformer-based methods such as CTformer \cite{wang2023ctformer}, hierarchical and/or hybrid designs including HCformer \cite{yuan2023hcformer}, FSformer \cite{kang2024fsformer}, and Swin-UNet \cite{cao2022swin}. However, supervised LDCT denoising methods depend heavily on paired normal-dose CT (NDCT) images, which are often scarce in clinical practice. 

To mitigate this data scarcity for supervision, alternative learning paradigms that relax or eliminate the need for clean references have been actively explored, including self-supervised learning approaches. Within the self-supervised family, a range of blind-spot denoising methods are attractive because they require only single noisy images and no assumption of paired noisy realizations, unlike Noise2Noise-style approaches. Representative methods include Noise2Self \cite{batson2019noise2self}, Noise2Void \cite{krull2019noise2void}, Noise2Same \cite{xie2020noise2same}. However, because the self-supervised loss is evaluated only at masked pixel locations, these methods leave a large portion of the available pixel-wise information underutilized in a single training step, limiting the proportion of pixels that directly contribute to the training objective at a given step, and the effective utilization of the CNN network's receptive field is constrained by mask sparsity. This motivates us to explore a progressive training strategy that distributes the supervision across multiple denoising steps.

We aim to address the under-utilization of pixel-wise supervisory information in blind-spot self-supervised denoising methods \cite{wu2020unpaired}, where supervision is typically restricted to masked locations within a single training step. Inspired by the observation of Zhu et al. \cite{zhu2025self} that the mapping from low-quality LDCT images to high-quality NDCT images can be progressively approximated by neural networks, we propose a progressive masked refinement learning scheme for LDCT denoising. Our method rests on three components. First, since no clean NDCT target is available and directly using the LDCT image as its own target would collapse to a trivial identity mapping, we inject independent, zero-mean Gaussian and Poisson noise into the input and target separately, forming a Noise2Noise training pair whose optimal solution under an $L_1$ loss recovers the underlying signal rather than the identity; the injected noise additionally regularizes training and mitigates overfitting. Second, we apply random masks to either the raw LDCT input or the intermediate denoised output. As demonstrated in Section \ref{sec:method}, masking can reduce the distribution gap between synthetic training data and LDCT. Third, we implement the denoising during training progressively across multiple stages, where different random masks are applied at different stages. This will alleviate the under-utilization of current blind-spot denoising methods that only calculate loss at blind spot areas. We treat these progressive learning stages for denoising as iteratively refining noisy training data to improve denoising stability. Extensive experiments in Section \ref{sec:exp} demonstrate that the proposed method achieves performance competitive with or superior state-of-the-art self-supervised denoising approaches and representative supervised methods.

\section{Related work}

\subsection{Self-supervised denoising model}
The self-supervised learning techniques, collectively known as Noise2- methods, have substantially advanced denoising performance. Noise2Noise \cite{lehtinen2018noise2noise} trains a network to map one noisy observation to another under the assumption that noise realizations are independent and zero-mean, thus requiring multiple noisy instances of the same image without clean targets. From a probabilistic perspective, Noise2Score \cite{kim2021noise2score} further relaxes these requirements by formulating self-supervised denoising as a Bayesian score estimation problem. In the context of computed tomography, Noise2Inverse \cite{hendriksen2020noise2inverse} extends the Noise2Noise paradigm by splitting a single raw sinogram into independent subsets of projection angles, producing multiple noisy reconstructions of the same anatomy. However, in clinical practice, obtaining two independent realizations of the same anatomy, each contaminated by different instances of zero-mean noise, is often infeasible due to radiation dose constraints and patient motion. Furthermore, Noise2Sim \cite{niu2022noise} represents a spatial relaxation of the Noise2Noise paradigm by leveraging structural similarities across intrinsically registered sub-images. Rather than requiring identical repeated scans, it utilizes adjacent CT slices or similar anatomical patches as noisy training pairs, effectively suppressing both independent and correlated noise by exploiting the natural redundancy in medical volumes. Despite the strong performance of Noise2Sim, their efficacy in Computed Tomography is often constrained by the underlying physical complexity of the noise field. Unlike the simplified additive noise models used in natural image denoising, the noise in LDCT is more accurately characterized by a compound Poisson distribution arising from polychromatic X-ray generation. Furthermore, the fundamental geometry of CT reconstruction creates a non-local noise dependency; a single point in the reconstructed spatial domain is mathematically linked to an entire sinusoid in the projection domain. Because these sinusoids intersect and overlap for different spatial points, the resulting noise is not only non-Gaussian but also highly correlated across the image, challenging the core 'independent noise' assumption \cite{wang2025bsn}. In contrast, Noise2Self \cite{batson2019noise2self} formalizes the concept of $\mathcal{J}$-invariant denoising, in which the prediction of each pixel is constructed to be independent of its own noisy observation. By enforcing this conditional independence through architectural constraints, effective denoising becomes possible using only single-sample noisy inputs, without clean targets or paired noisy acquisitions. Building on this framework, Noise2Void \cite{krull2019noise2void} further introduces a masking strategy in which selected pixels are predicted from their surrounding context, relying on the assumptions that noise is independent across pixels and that the underlying signal can be inferred from neighboring pixels. Meanwhile, Neighbor2Neighbor \cite{li2019low} avoids explicit masking and instead uses that neighboring patches share the same underlying clean signal for denoising. Additionally, Noise2Same \cite{xie2020noise2same} addresses the information loss introduced by a consistency constraint between masked and unmasked predictions, such as Noise2Void. The Similarity-based Visual Blind-spot (SVB) scheme \cite{wang2024svb} extends the Noise2Void framework to LDCT denoising by integrating structural redundancy from similar anatomical patches with a specialized masking strategy. This approach improves the utilization of self-supervised information by combining similarity-based visual correspondence with a blind-spot scheme.

\subsection{Denoising diffusion probabilistic model}
Beyond blind-spot denoising, diffusion-based approaches have recently attracted increasing attention for LDCT denoising due to their ability to progressively refine degraded images toward high-quality anatomical representations. Diffusion Probabilistic Models (DPMs) \cite{ho2020denoising} learn a prior distribution of high-quality images and formulate denoising as an iterative inverse problem, where the observed LDCT image guides the reverse process toward a clean state \cite{gao2025noise, Zhang2024PartitionedHDA, liu2025diffusion}. Recent studies have explored diffusion-based priors to alleviate the reliance on paired LDCT-NDCT training data. Zhang et al. \cite{Zhang2024PartitionedHDA} proposed Partitioned Hankel-based Diffusion (PHD), a few-shot unsupervised framework that exploits the low-rank structure of projection-domain Hankel matrices and combines a diffusion prior with iterative reconstruction to recover high-quality CT images from limited training data. Liu et al. \cite{liu2025diffusion} introduced Dn-Dp, a zero-shot denoising framework that learns diffusion priors exclusively from NDCT images and incorporates the observed LDCT image as a likelihood term during the reverse diffusion process. Adaptive coefficients are further introduced to balance the learned prior and measurement fidelity. More recently, Gao et al. \cite{gao2025noise} proposed NEED, which incorporates the physical noise characteristics of CT acquisition into the diffusion process through a shifted Poisson diffusion model in the projection domain and a doubly guided diffusion model in the image domain. A timestep-matching strategy further enables the model to adapt to different and unseen dose levels. While these approaches demonstrate the effectiveness of diffusion priors for LDCT denoising without requiring paired LDCT-NDCT supervision, they primarily rely on a pretrained generative prior to constrain the reconstruction process. In contrast, the self-supervised denoising paradigm learns directly from LDCT observations, without requiring either paired clean targets or an externally pretrained NDCT diffusion prior.

\section{Methodology}
\label{sec:method}

\subsection{Limitation of Noise2Self}
We first consider LDCT denoising in a Noise2Self learning setting. Given a LDCT image $x_i$ and the corresponding NDCT image $y_i$, assuming $x=x_c+n_0$, where $x_c$ is the clean CT images and $n_0$ is the additive noise. Assuming that noise $n_0$ is zero-mean and independent across all dimensions, self-supervised learning methods such as Noise2Self can be formulated.

Noise2Self is trained with the loss function $\mathcal{L}(f)=E_x||f_{J}(\mathbf{x}_{J^{c}})-\mathbf{x}_{J}||^2$. $J^{c}$ is the complement set of $J$, the blind spot areas. Importantly, the function $f$ is required to be $\mathcal{J}$-invariant, as established by Batson \textit{et al.} \cite{batson2019noise2self} to prevent learning the identity mapping with a random masking. Thanks to $\mathcal{J}$-invariant and the zero-mean assumption of noise, minimizing the self-supervised loss $E_x||f_{\theta}(\mathbf{x})-\mathbf{x}||^2$ indirectly minimizes the supervised loss $E_x||f_{\theta}(\mathbf{x})-\mathbf{y}||^2$ \cite{xie2020noise2same}, which can be expressed as:
\begin{equation}
E_x||f_{\theta}(\mathbf{x})-\mathbf{x}||^2 \propto E_x||f_{\theta}(\mathbf{x})-\mathbf{y}||^2
\label{J_invariant}
\end{equation}
where, $f_\theta$ represents denoising networks with optimal parameters $\theta$, and $L$ denotes the loss function. Current implementations for computing the $\mathcal{J}$-invariant denoising function $f$ rely on blind-spot networks \cite{krull2019noise2void, batson2019noise2self, laine2019high}. However, previous blind-spot denoising methods may not fully utilize the available supervisory information beyond the masked locations. Our objective is therefore to solve the problem of blind-spot networks, enabling the network to progressively recover from LDCT images. This motivates the progressive masked refinement learning introduced in Section \ref{j-learning}.

\subsection{Progressive Masked Refinement Learning}
\label{j-learning}

Since paired LDCT–NDCT images are generally unavailable in practical settings, directly supervised training a denoising network with NDCT targets is often infeasible. To enable self-supervised training, we construct training pairs from a single LDCT observation by adding two independently sampled noises: we add independent controlled Gaussian and Poisson noise to the same LDCT image, one perturbed version serves as the network input and the other as the target. This strategy is valid under the assumption that the signal intensity dominates the noise, i.e., $\mathbb{E}[\mathbf{x}]\gg\mathbb{E}[\mathbf{n}_0]$ and $\mathrm{Var}[\mathbf{x}]\gg\mathrm{Var}[\mathbf{n}_0]$, so that the LDCT observation is a close approximation of the underlying clean image; this assumption has been shown to hold for LDCT denoising \cite{zhu2025self}. 

However, the added corruption introduces a distributional discrepancy between the synthetic training inputs and the original LDCT observations seen at inference. We therefore investigate whether random masking can reduce this discrepancy by restricting the pixels available to the network. To establish a theoretical basis, we compare the Kullback–Leibler (KL) divergence between the two distributions with and without masking.

Let $p$ and $q$ denote the distributions of the synthetic (noise-added) and original LDCT observations, and let $\mathbf{a}^{s}=M\odot\mathbf{x}^{s}$, $\mathbf{b}^{s}=(1-M)\odot\mathbf{x}^{s}$ be the visible and masked pixels. Since masking discards $\mathbf{b}$, the network only sees the marginal over visible pixels $\mathbf{a}$. By the chain rule of KL divergence, the divergence over visible pixels lower-bounds that over full images \cite{chihaoui2025unsupervised}:

\begin{align*}
D_{KL}\big(p(\mathbf{a},\mathbf{b})\,\|\,q(\mathbf{a},\mathbf{b})\big)
&=\int p(\mathbf{a})\,
D_{KL}\big(p(\mathbf{b}|\mathbf{a})\,\|\,q(\mathbf{b}|\mathbf{a})\big)\,
d\mathbf{a}
+D_{KL}\big(p(\mathbf{a})\,\|\,q(\mathbf{a})\big)\\
&\ge D_{KL}\big(p(\mathbf{a})\,\|\,q(\mathbf{a})\big)
\label{KL_mask}
\end{align*}
which holds since $p(\mathbf{a})\ge 0$ and
$D_{KL}\big(p(\mathbf{b}|\mathbf{a})\,\|\,q(\mathbf{b}|\mathbf{a})\big)\ge 0$.
Thus, masking the inputs reduces the divergence between the synthetic
training distribution and the original LDCT distribution, narrowing the
train–inference gap.

However, recovering a clean estimate from a heavily masked input in a single step is difficult, as each mask discards informative pixels. Rather than denoising in one step, we refine the estimate progressively: the output of each step becomes the input to the next, allowing masked content to be recovered incrementally. Consider a temporally progressive denoising process where $\hat{x}^t$ denotes the denoised LDCT image at time step $t$ and $noise(t)$ represents the corresponding subject noise w.r.t. denoising function $f$, given that $f(\hat{x}^t) = \hat{x}^t - noise(t)$ after denoising. Meanwhile, the denoised image $\hat{x}^t$ at time step $t$ serves as the input for the next time step $t+1$, such that $f(\hat{x}^t)=\hat{x}^{t+1}$. Intuitively, each additional step allows 
the network to aggregate more contextual information and further refine 
the estimate, motivating a progressive refinement scheme. Then, the denoising process with a total of $K$ time steps is illustrated as follows:
\begin{align*}
\hat{x}^t=f(\hat{x}^t)+noise(t) \\
\hat{x}^{t+1}=f(\hat{x}^{t+1})+noise(t+1) \\
...\\
\hat{x}^{t+K}=f(\hat{x}^{t+K})+noise(t+K)
\end{align*}

Progressive refinement plays two roles. First, it decomposes the
difficult single-step mapping into a sequence of easier incremental
refinements, where masked regions are filled by aggregating context over
successive steps. Second, each intermediate estimate, paired with a newly
sampled mask, serves as an additional training sample, expanding a single
LDCT image into $K$ progressively refined training pairs and alleviating
the sparse-supervision problem of single-step masked training. In practice, rather than assuming convergence to a fixed point, we treat these intermediate outputs $\{\hat{x}^{t}, \hat{x}^{t+1}, \dots\}$ as progressively refined training samples and use a finite number of steps $K$.

Therefore, the final loss function at a given time step $t$ is defined as follows:
\begin{equation}
   \underset{\theta_{t}}{argmin} \sum_i L( f_{\theta_{t}}(M\odot (\hat{x}_i^t+ \mathds{1}_{\{\hat{x}_i^t = x\}} n_1), (x_i+n_2))
\label{self_loss}
\end{equation}

Where $M\sim \mathcal{M}_{\alpha}\{ M_0,M_1,...,M_K\}$ is a binary random mask. $\mathcal{M}_\alpha$, $\alpha \in (0,1)$ denotes a set of masks with random sampling rate $\alpha$. $\odot$ represents the Hadamard product (i.e., element-wise product) \cite{horn1990hadamard}. $\mathds{1}_{\{\hat{x}_i^t = x\}}$ is the indicator function that equals one when the input is an LDCT image. Similar to supervised temporal denoising, the update is given by $\hat{x}_i^{t+1}=f_\theta(M\odot \hat{x}_i^t)$.

In this work, we generate $n_1$ and $n_2$ using a combination of zero-mean additive white Gaussian noise (AWGN), sampled from the distribution $\mathcal{N}(0,\sigma^2)$, and Poisson noise, a noise model validated in \cite{zhu2025self}.

\begin{figure}[!h]
\centering
\includegraphics[width=0.8\textwidth]{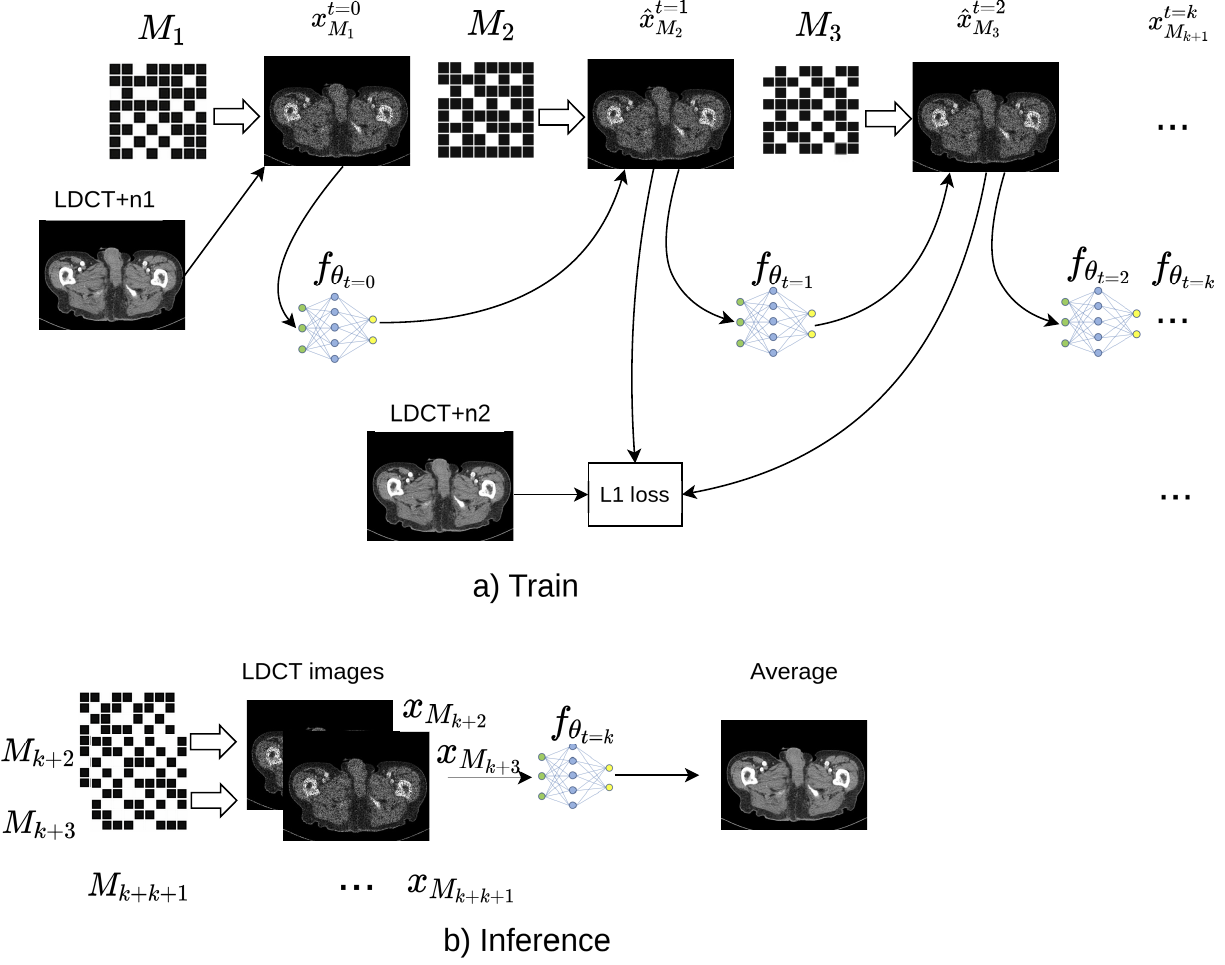}
\caption{Overview of the progressive masked refinement learning. a) During training, a random mask is applied to the LDCT image at each time step. $\hat{x}_{M_i}^{t}$ is the masked denoised image $x_{M_i}$ at time $t$.  b) During inference, multiple randomly masked versions of the same LDCT image are fed into the trained model, and the final denoised result is obtained by averaging the corresponding outputs. }\label{fig1}
\end{figure}

Concretely, at each step $t$, a random mask $M_{\alpha,i,t}$ is applied and the model is optimized by minimizing Eq. (\ref{self_loss}). Note that the
model evolves across time steps during training, whereas at inference a
single converged model is applied at every step; the progressive training
thus shapes one model to refine consistently at all steps, rather than
learning a separate model per step.  At inference, a high masking ratio inevitably discards structured information, so the reconstruction from any single mask carries uncertainty in the masked regions. To ensure robust reconstruction, we adopt a prediction ensemble: we sample $k$ differently masked versions of the LDCT image and average the corresponding outputs. By aggregating
complementary contextual cues from different sets of visible pixels, the ensemble exploits the inherent redundancy and local correlation of images, mitigating the reconstruction uncertainty of individual masks and yielding a more accurate, detailed output. An overview of the proposed progressive masked refinement learning is shown in Fig. \ref{fig1}. Unlike diffusion models, our method does not rely on a predefined stochastic diffusion process. Furthermore, our approach removes the requirement for NDCT images, making it highly applicable to practical scenarios where paired data is unavailable.

\section{Experiments}
\label{sec:exp}

We first introduce the configuration of dataset and experiments. Then, we evaluate the contribution of (a) the progressive masked refinement learning w.r.t. time step $k$, (b) noise injection configurations w.r.t. noise level $\sigma$, (c) mask ratio $\alpha$, and (d) the noise combination with four ablation experiments. Finally, we compare our method to several supervised and self-supervised methods under two conditions: sample-wise comparison on the LDCT dataset and patient-wise evaluation. 

\subsection{Experimental setup}
We conduct our experiments on the Low-Dose CT Grand Challenge dataset\footnote{\url{https://www.aapm.org/grandchallenge/lowdosect/}}, a publicly available benchmark that provides paired low-dose and standard-dose CT scans acquired under identical imaging conditions \cite{moen2021low}. The dataset includes scans from multiple anatomical regions, such as abdomen sub-dataset and chest sub-dataset. Each volume is reconstructed with an in-plane resolution of $512 \times 512$ and a voxel spacing of $0.5859 \times 0.5859 \times 3.0 mm^{3}$. A total of 36 patients are included, with 27 used for training (20) and validation (7), and the remaining 9 reserved for evaluation to ensure subject-level separation. All images are provided in DICOM format and are first converted to Hounsfield Units (HU) to preserve the physical meaning of CT attenuation values. The intensities are then normalized to the range $[0,1]$ using a fixed HU window of $[-1024,3072]$, which covers air, soft tissue, and high-density structures while maintaining consistency across subjects and experiments. Note that our method does not require any NDCT at the training stage.

We implement our model using the PyTorch 2.7 library and train it on a server equipped with a single NVIDIA V100 GPU. The Adam optimizer is employed with parameters $\beta_1 = 0.9$, $\beta_2 = 0.99$, and $\epsilon = 10^{-8}$. Model parameters are optimized using the $L1$ loss. The initial learning rate was set to $10^{-3}$ and decayed by a factor of two every 20 epochs. Training was conducted for a total of 100 epochs. We use a batch size of 4, with 10 patches per image and a patch size of $128 \times 128$. 

For quantitative assessment, model performance is evaluated using three widely adopted image quality metrics: peak signal-to-noise ratio (PSNR), structural similarity index (SSIM), and root mean squared error (RMSE). These metrics are computed in accordance with the evaluation protocol described in Chen \textit{et al.} \cite{chen2017low}, enabling consistent and reproducible comparison across methods. PSNR and RMSE quantify pixel-wise fidelity between the denoised and reference images, while SSIM measures structural and perceptual similarity. The mathematical formulations of PSNR and SSIM are shows in Eq. \ref{psnr} and Eq. \ref{ssim} respectively. We use the standard deviation as a measure of variability to assess the stability of the models.

\begin{equation}
    PSNR=10log_{10}(\frac{MAX^2}{\frac{1}{n}\sum^n_{i=1}(m_i-l_i)^2})
    \label{psnr}
\end{equation}
where $m_i$ and $l_i$ are pixels of LDCT or denoised CT images and NDCT images, respectively. n is the number of pixels in the image and $MAX$ denotes the maximum value of the image pixels.

\begin{equation}
    SSIM(m,l)=\frac{(2\mu_m\mu_l+c_1)(2\sigma_{ml}+c_2)}{(\mu_m^2+\mu_l^2+c_1)(\sigma_{m}^2+\sigma_{l}^2+c_2)}
    \label{ssim}
\end{equation}
where $\mu_m$ and $\mu_l$ are the means of $m$ and $l$, respectively; $\sigma_{m}^2$ and $\sigma_{l}^2$ are variances of $m$ and $l$, respectively. $\sigma_{ml}$ is the covariance of $m$ and $l$, and $c_1$ and $c_2$ are two hyperparameters for stabilizing the division operation.

\subsection{Method comparison}
\subsubsection{Comparison Across Samples on Abdominal and Chest Datasets}
We compared our method with several representative supervised and unsupervised approaches, including RED-CNN \cite{chen2017low}, Swin-Unet \cite{cao2022swin}, CycleGAN \cite{zhu2017unpaired}, BM3D \cite{dabov2007image}, Noise2Void \cite{krull2019noise2void}, and Neighbor2Neighbor \cite{huang2021neighbor2neighbor}. Among these, RED-CNN, Swin-Unet, and CycleGAN were trained in a supervised manner, whereas the remaining methods were trained solely on LDCT images. Except for BM3D, RED-CNN, and Swin-Unet, other baselines adopt a U-Net–based backbone. It is worth noting that, unlike Noise2Self, we operate on a standard U-Net architecture with batch normalization instead of instance normalization.

Table \ref{comparison} reports the quantitative comparison on the LDCT dataset with $k=5$, $\sigma=10$ and $\alpha=0.1$. It is observed that our method outperforms popular self-supervised denoising approaches across all evaluation metrics. In particular, the proposed progressive masked refinement learning with U-Net not only surpasses unsupervised baselines but also outperforms some supervised methods, such as Cycle-GAN, and achieves performance comparable to the widely adopted supervised RED-CNN.
Similar performance gains are observed when integrating the proposed method with Swin-Unet and RED-CNN backbones. Our models exhibit performance close to their corresponding supervised counterparts, demonstrating that the proposed strategy can be applied across different architectures and can effectively enhance denoising performance without introducing additional model complexity. We also evaluated our method on the Mayo Chest dataset, which presents a more significant challenge. As demonstrated in Table \ref{comparison_chest}, the performance is broadly consistent  with our previous findings. For this dataset, the optimal hyperparameters were configured as $k=2$, $\sigma=50$ and $\alpha=0.01$.

\begin{table}[t]
\centering
\begin{tabular}{l c c cc}
\cline{1-5}
Methods& PSNR &SSIM & RMSE  & parameters\\
\cline{1-5}
LDCT (Dataset) & 28.842 & 0.856 & 15.160  &-\\
\cline{1-5}
\multicolumn{5}{c}{\textit{Supervised}}         \\ \hdashline
RED-CNN & \textbf{31.748} & \textbf{0.896} & \textbf{10.686}& 1.85M\\
Swin-Unet \cite{cao2022swin} & 31.060&0.890& 11.558&  0.95M\\
Cycle GAN & 30.289& 0.879 & 12.813& 114M\\
\cline{1-5}
\multicolumn{5}{c}{\textit{Self-supervised}}         \\ \hdashline 
BM3D&29.103&0.842&14.281&-\\
Noise2Void& 29.489 &0.864& 14.014&31M\\
Neighbor2Neighbor& 30.078 &0.876& 13.099&31M\\
PMRL+Unet &\textbf{31.510} & \textbf{0.892}&\textbf{10.993} &31M\\
PMRL+Swin-Unet &30.570 & 0.881&12.134 &0.95M\\
PMRL+RED-CNN &31.261 & 0.889&11.324 & 1.85M\\
\cline{1-5}
\end{tabular}
\caption{Quantitative evaluation for LDCT abdomen dataset. PMRL denotes the proposed Progressive Masked Refinement Self-supervised learning. Note: The Swin-Unet model used in our experiments is a reduced version of the original model \cite{cao2022swin}, with fewer layers due to the adoption of a patch-based input strategy}\label{comparison}
\end{table}

\begin{figure}[H]
	\centering
	\subfigure[]{
		\includegraphics[width=0.3\textwidth,trim=50 10 80 10, clip]{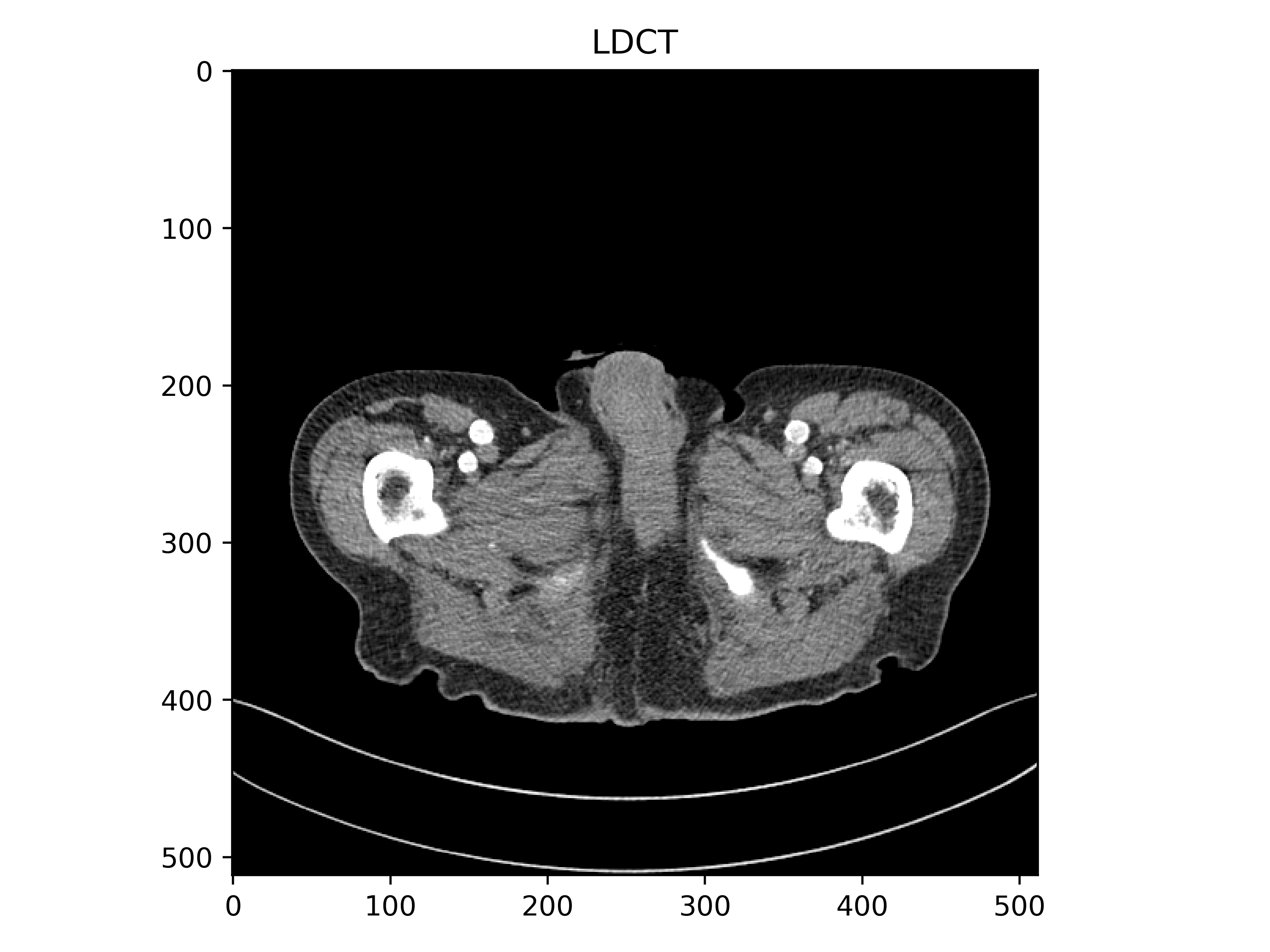}
	}
	\subfigure[]{
		\includegraphics[width=0.3\textwidth,trim=50 10 80 10, clip]{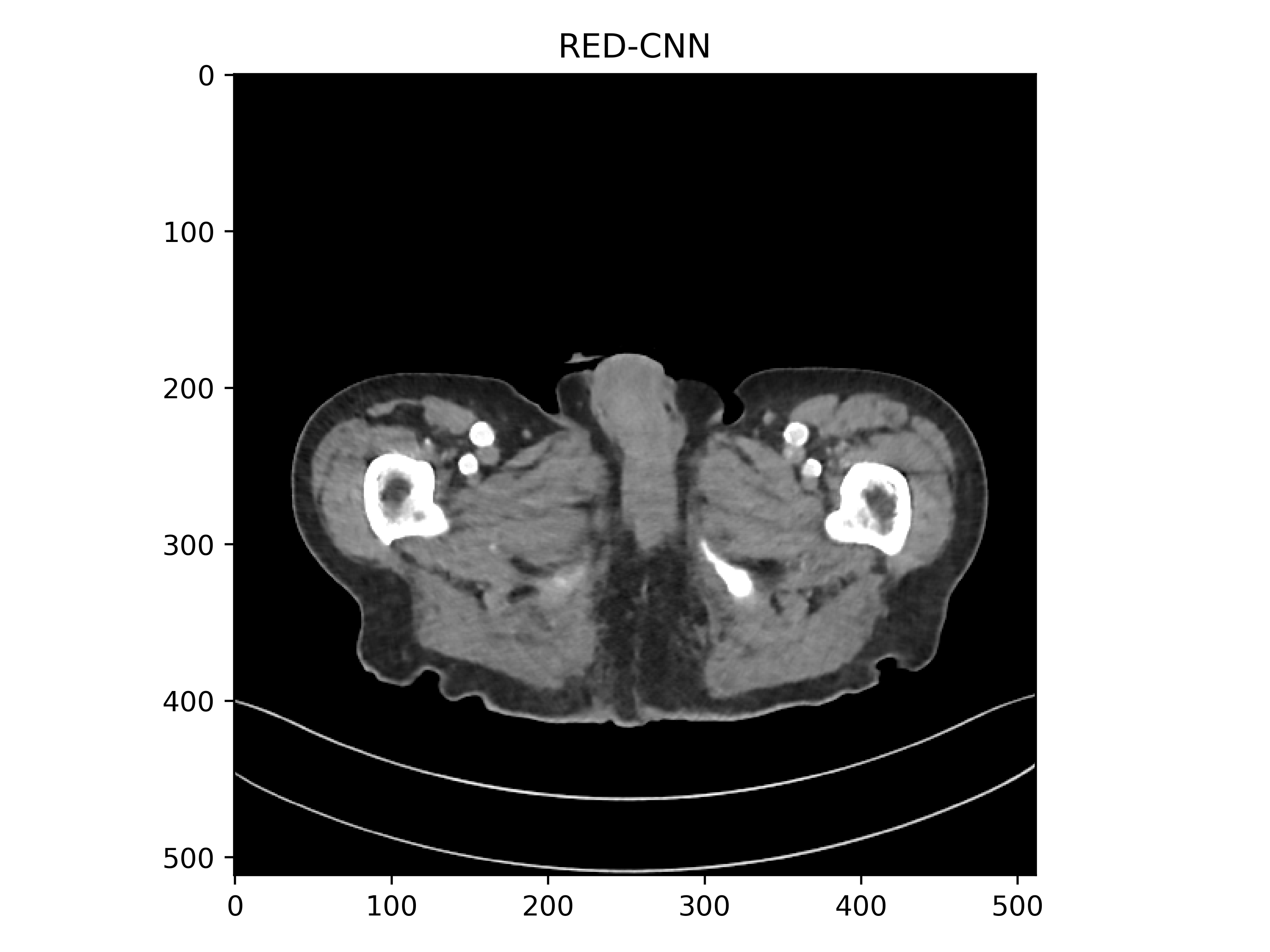}
	}
	\subfigure[]{
		\includegraphics[width=0.3\textwidth,trim=50 10 80 10, clip]{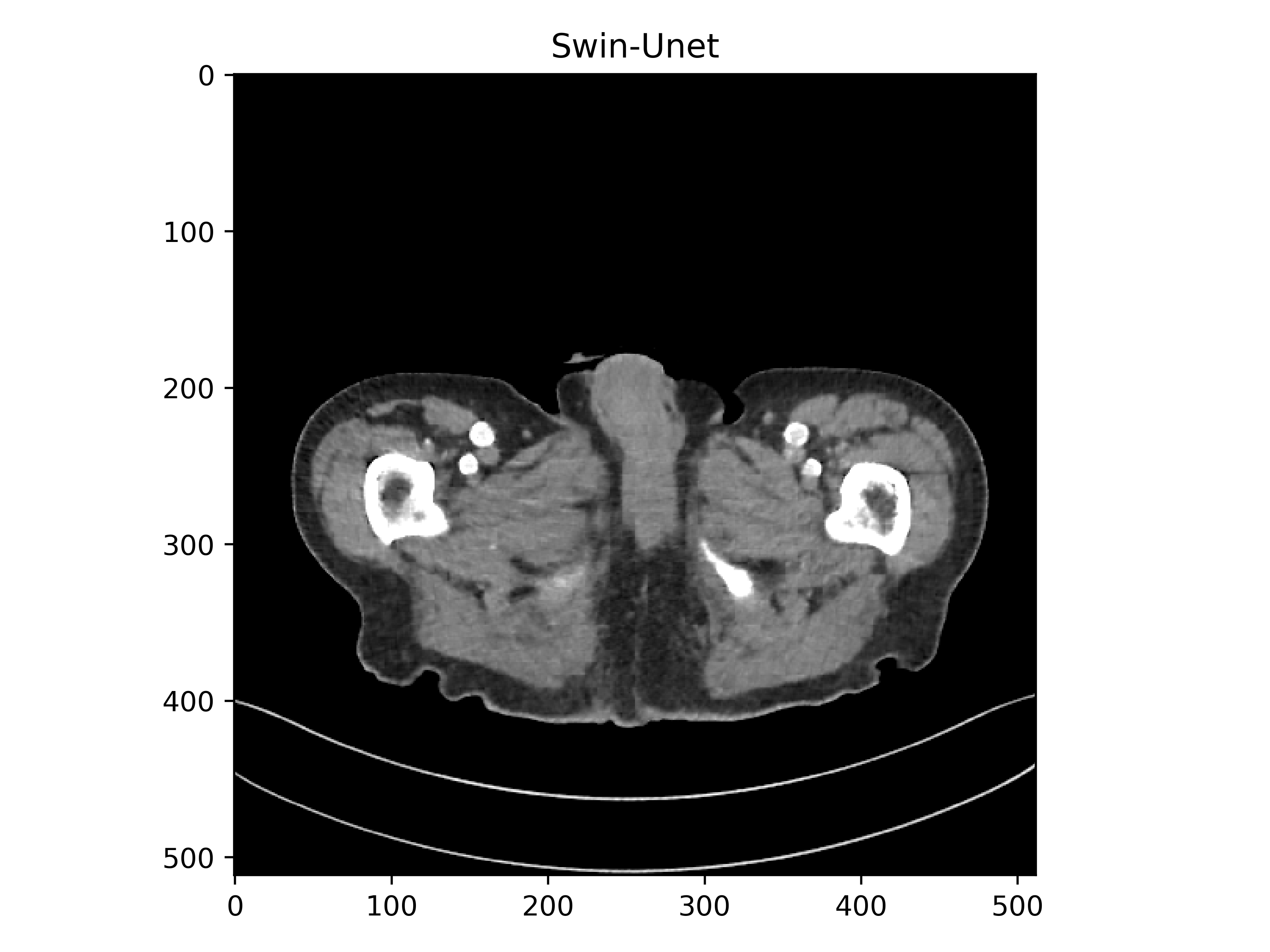}
	}
	\quad
	\subfigure[]{
		\includegraphics[width=0.3\textwidth,trim=50 10 80 10, clip]{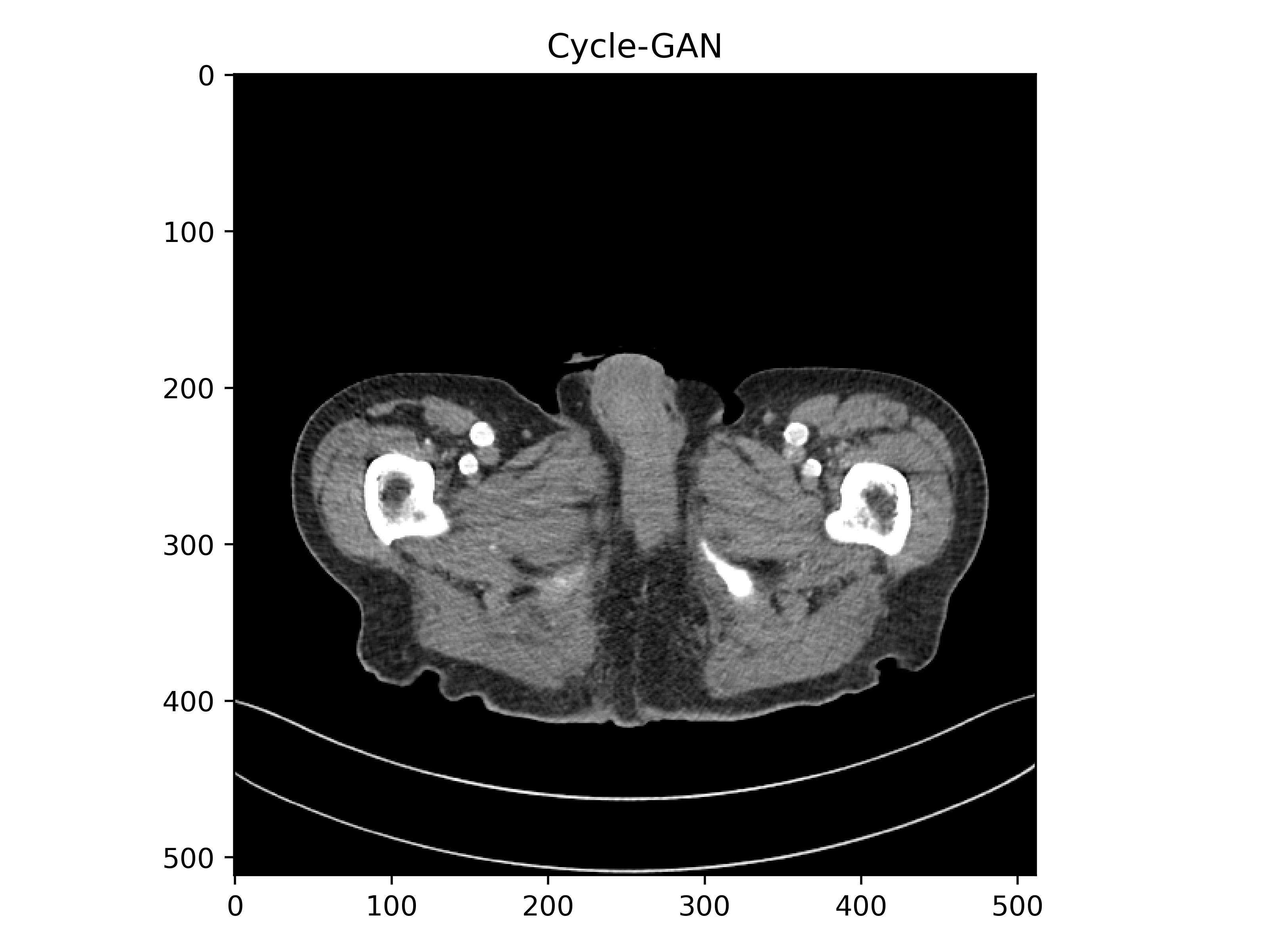}
	}
	\subfigure[]{
		\includegraphics[width=0.3\textwidth,trim=50 10 80 10, clip]{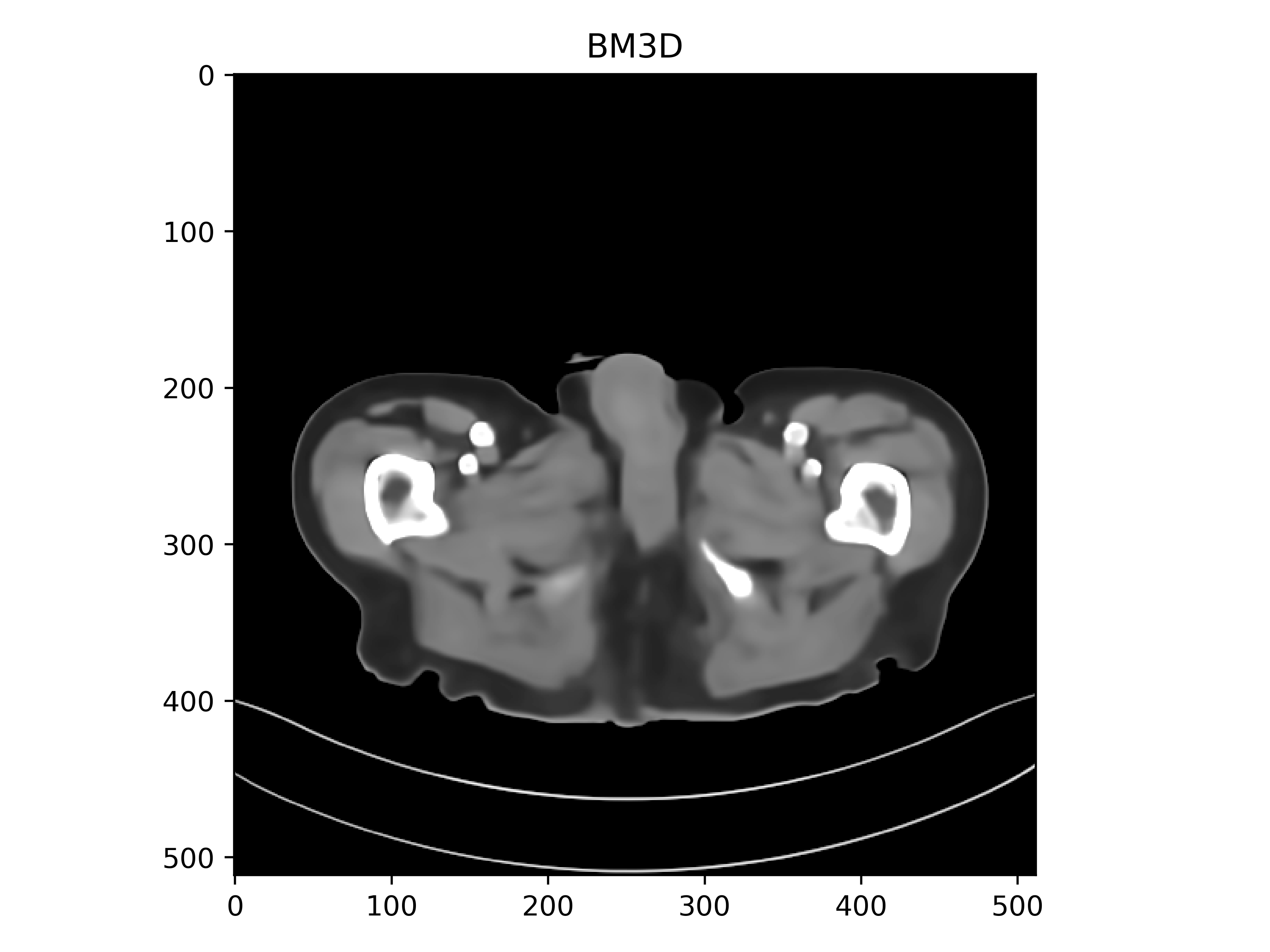}
	}
	\subfigure[]{
		\includegraphics[width=0.3\textwidth,trim=50 10 80 10, clip]{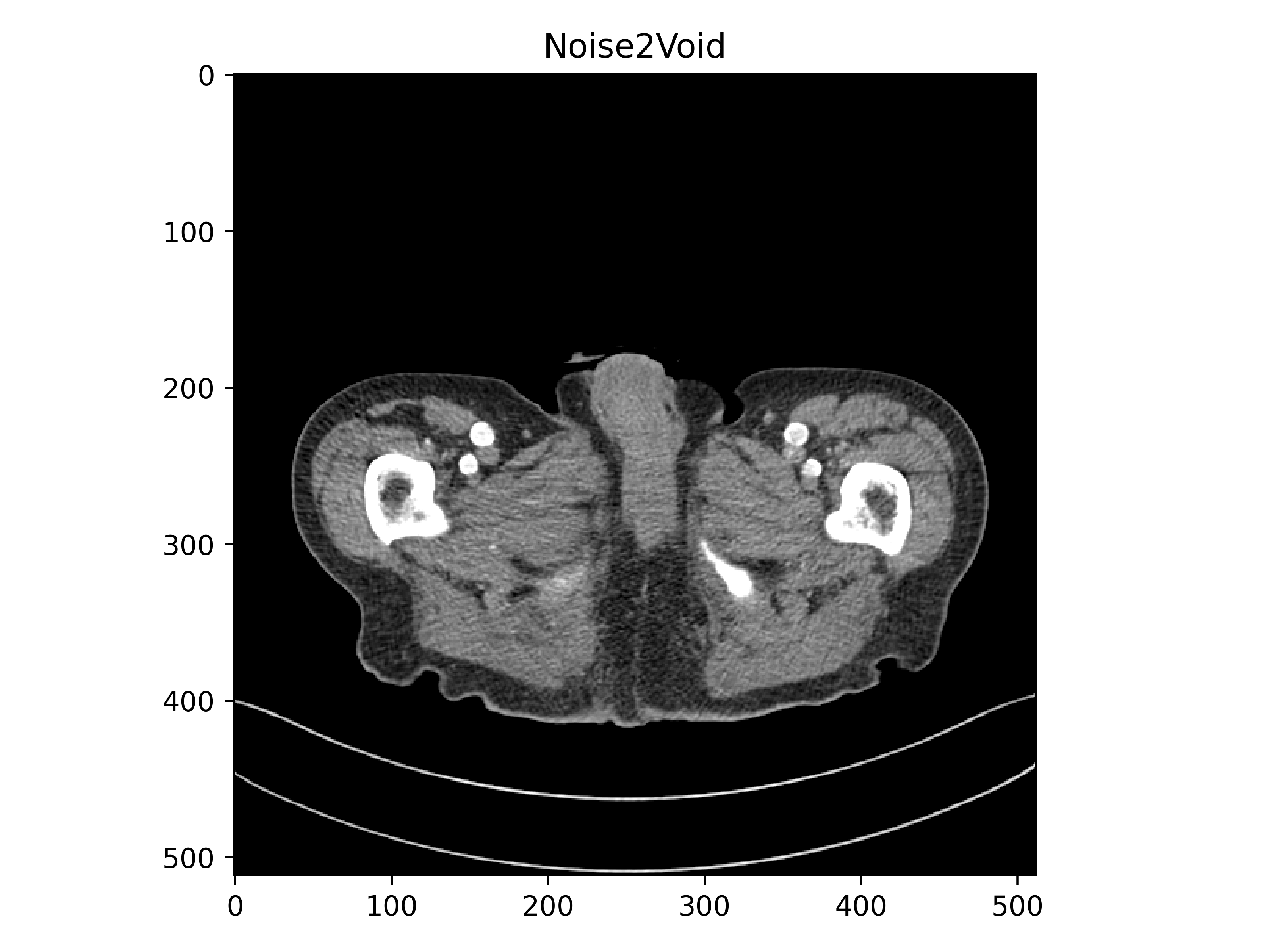}
	}
	\quad
	\subfigure[]{
		\includegraphics[width=0.3\textwidth,trim=50 10 80 10, clip]{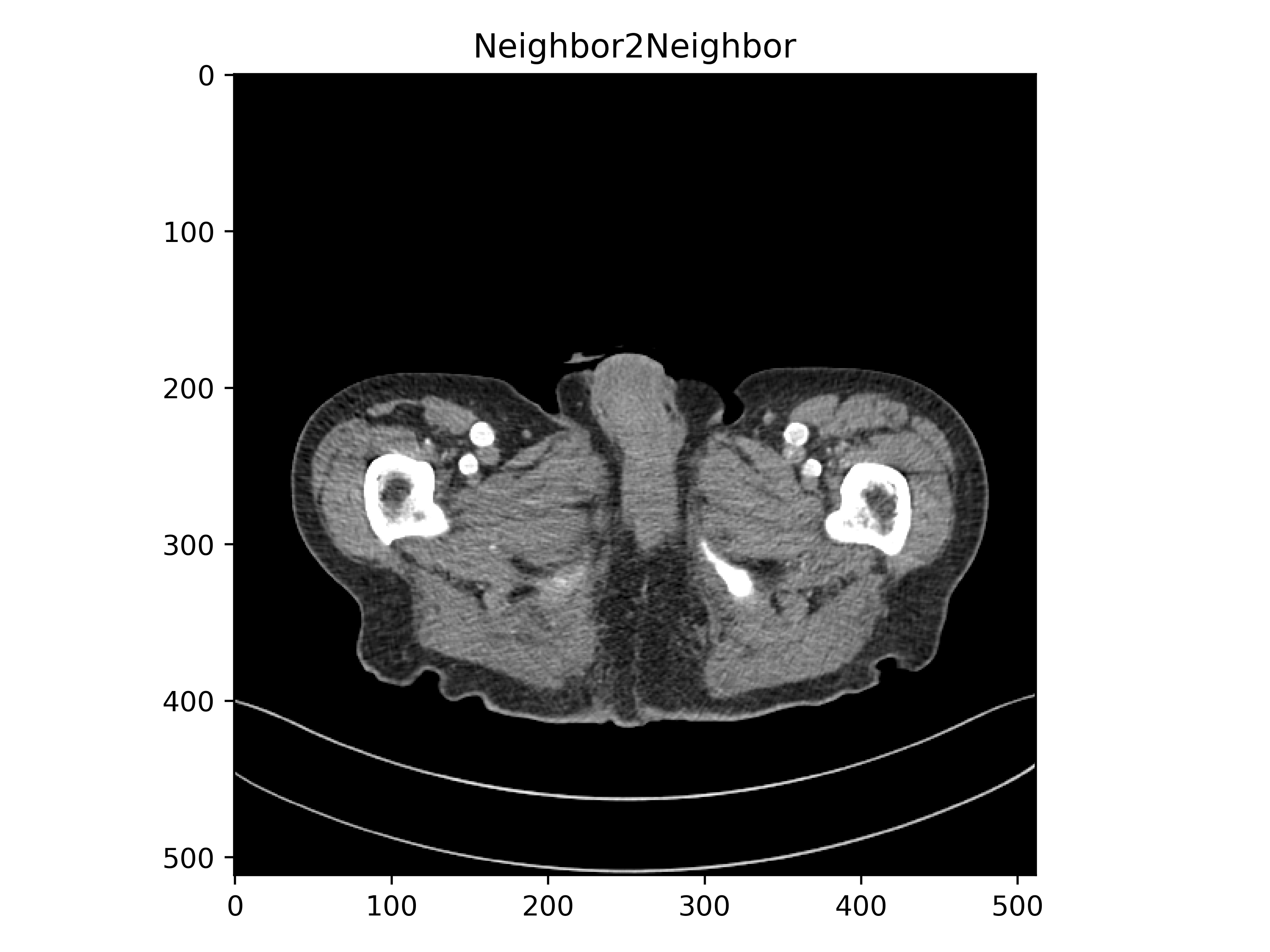}
	}
	\subfigure[]{
		\includegraphics[width=0.3\textwidth,trim=50 10 80 10, clip]{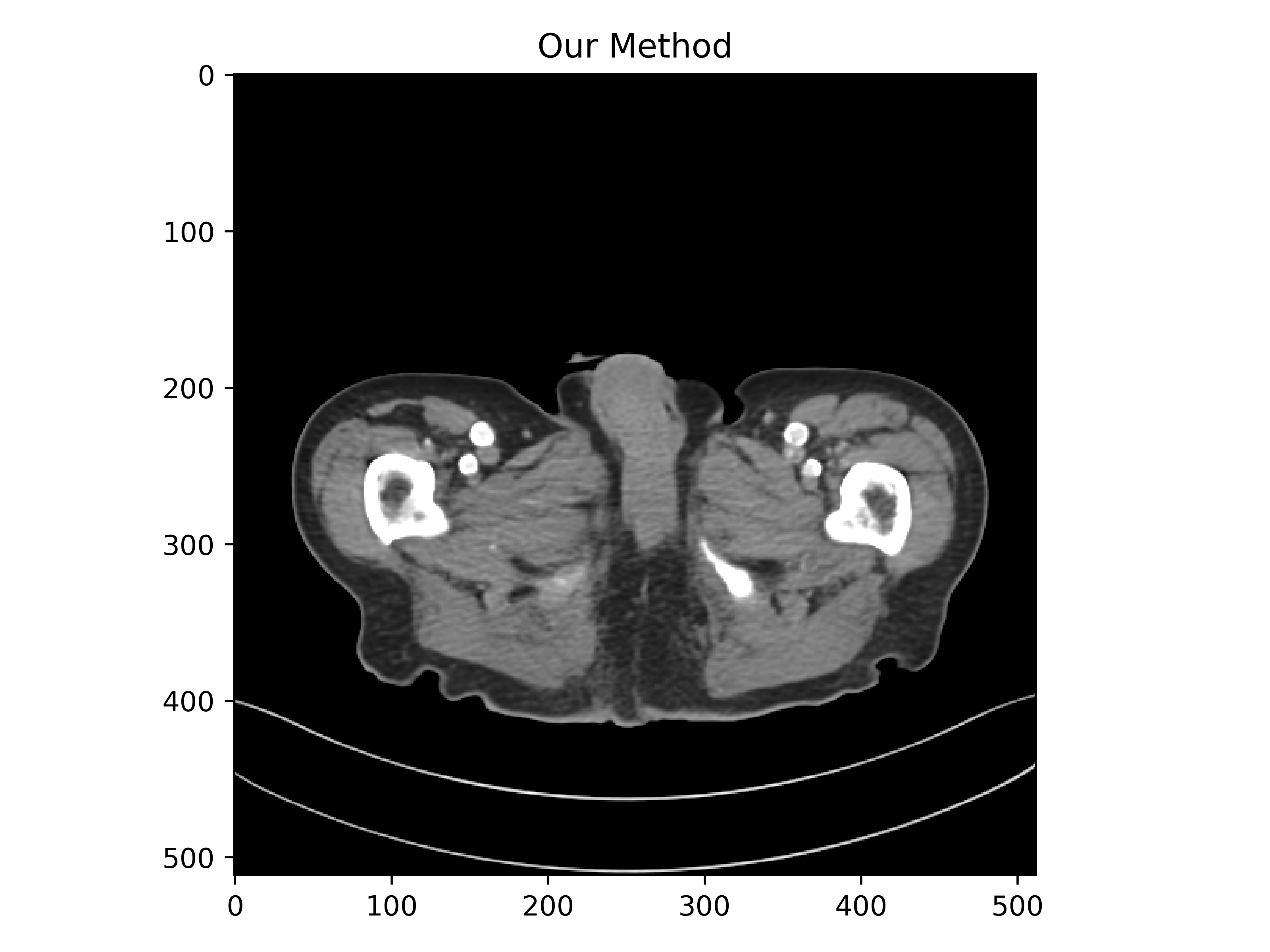}
	}
	\subfigure[]{
		\includegraphics[width=0.3\textwidth,trim=50 10 80 10, clip]{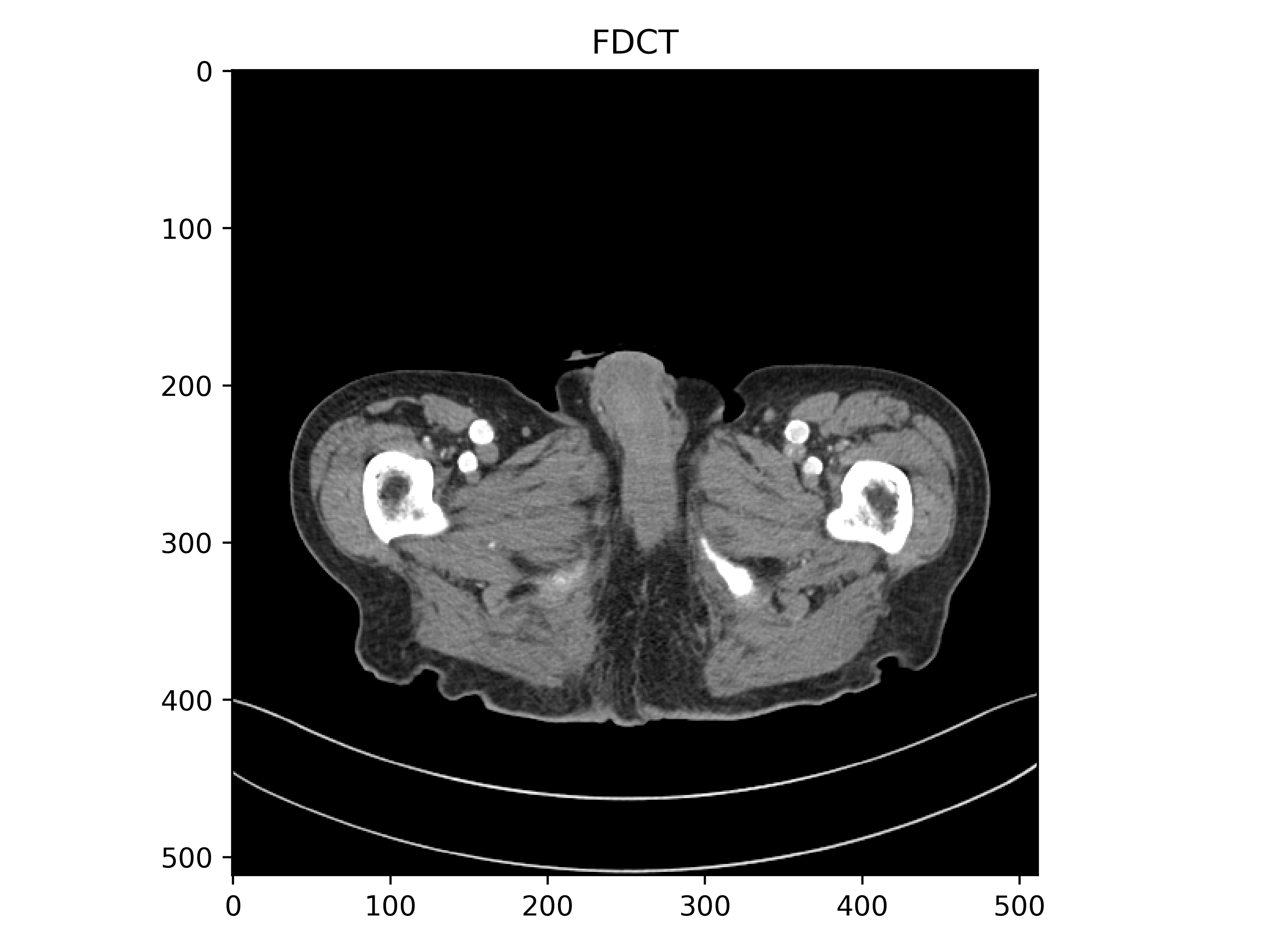}
	}
	
	\caption{Results of pelvis image for comparison w.r.t. NDCT. (a)LDCT$:$ 30.702dB, (b)RED-CNN$:$ 34.131dB, (c)Swin-Unet$:$ 33.416dB, (d)Cycle-GAN$:$ 32.761dB, (e)BM3D$:$ 31.214dB, (f)Noise2Void$:$ 31.536dB, (g)Neighbor2Neighbor$:$ 32.125dB, (h)our PMRL$+$Unet$:$ 33.684dB, (I)Full-dose (FD)CT/NDCT.}
	\label{LDCT_fig}
\end{figure}

\begin{figure}[t]
	\centering
	\subfigure[]{
		\includegraphics[width=0.3\textwidth,trim=50 10 80 10, clip]{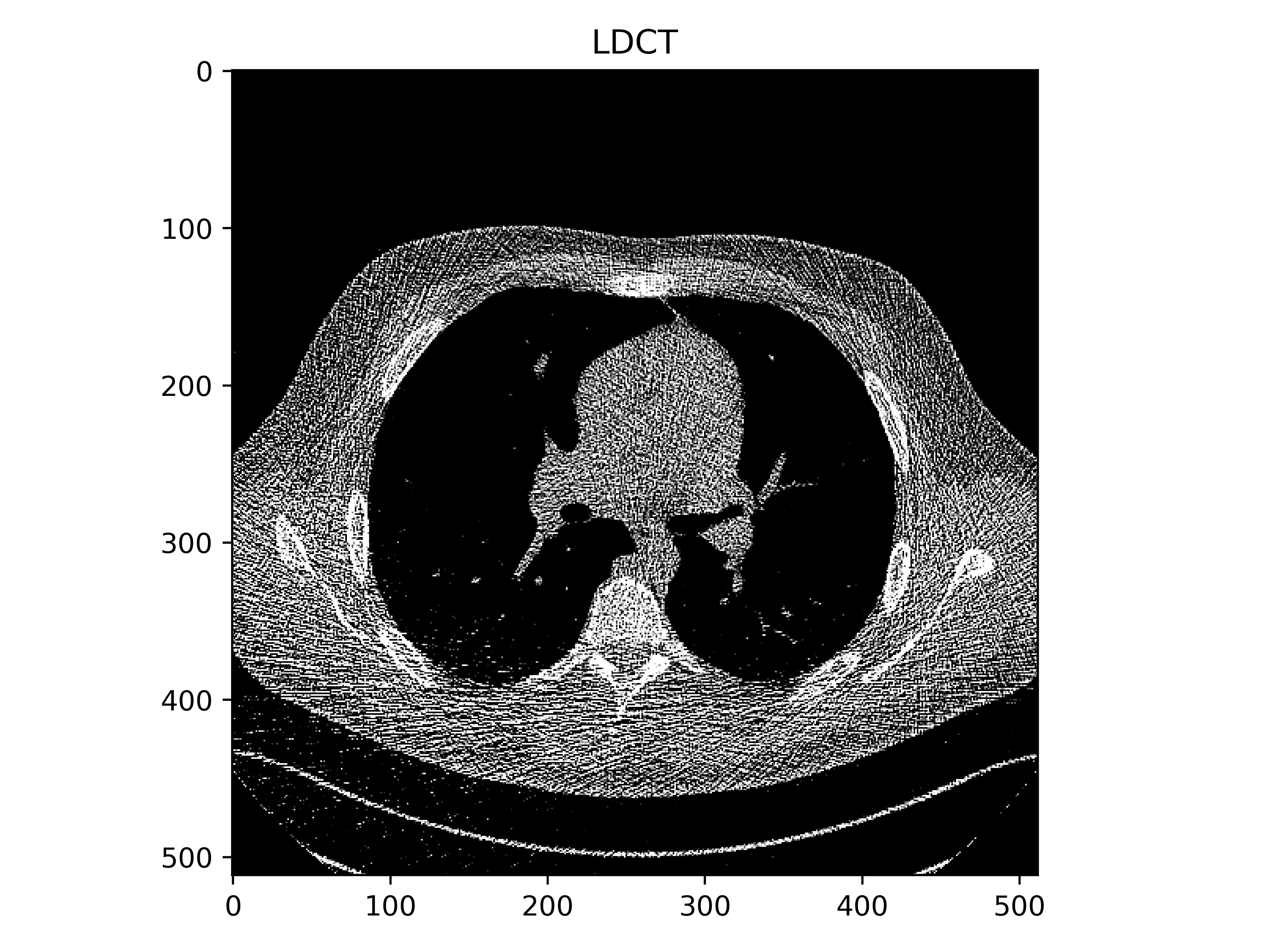}
	}
	\subfigure[]{
		\includegraphics[width=0.3\textwidth,trim=50 10 80 10, clip]{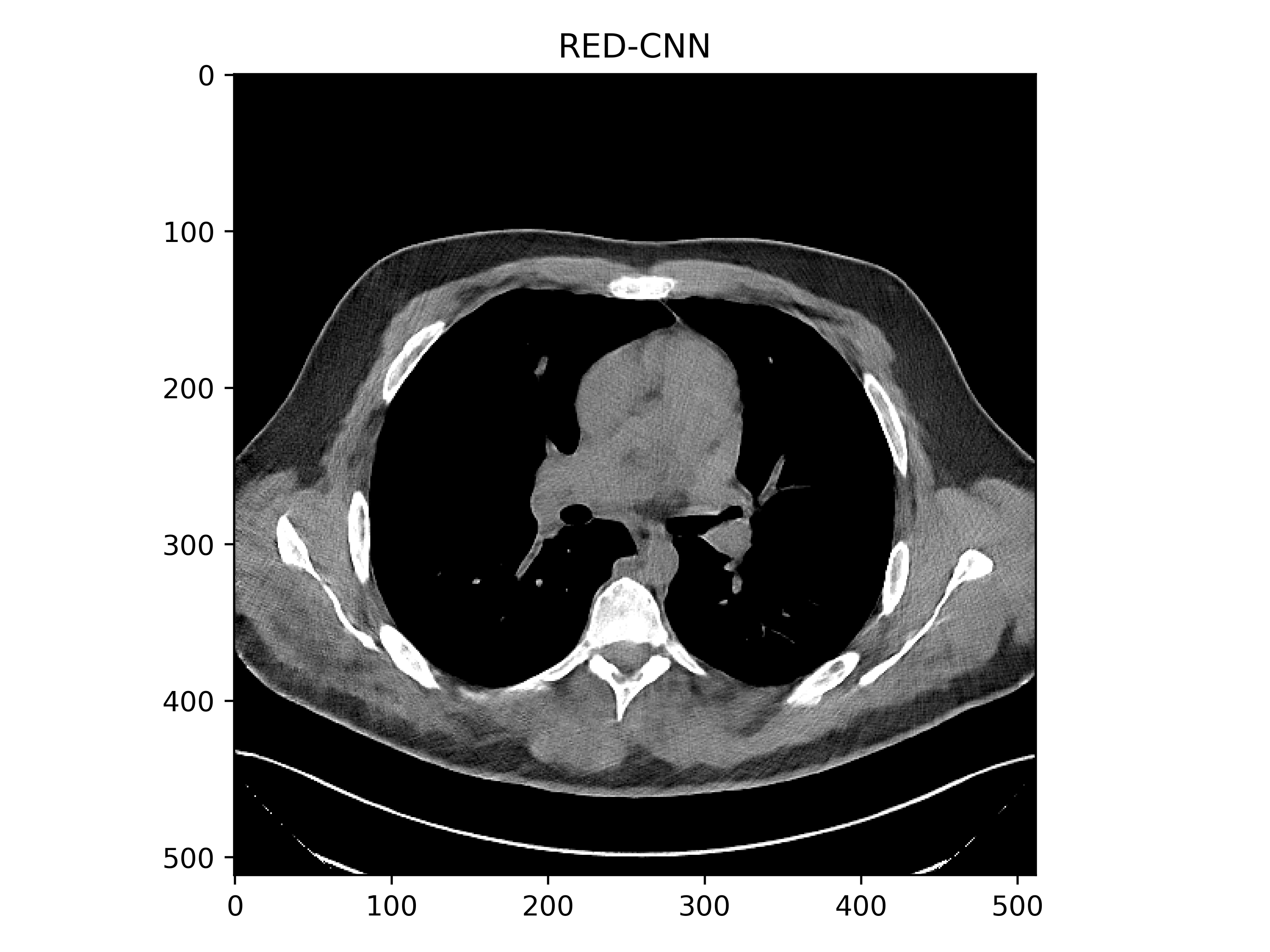}
	}
	\subfigure[]{
		\includegraphics[width=0.3\textwidth,trim=50 10 80 10, clip]{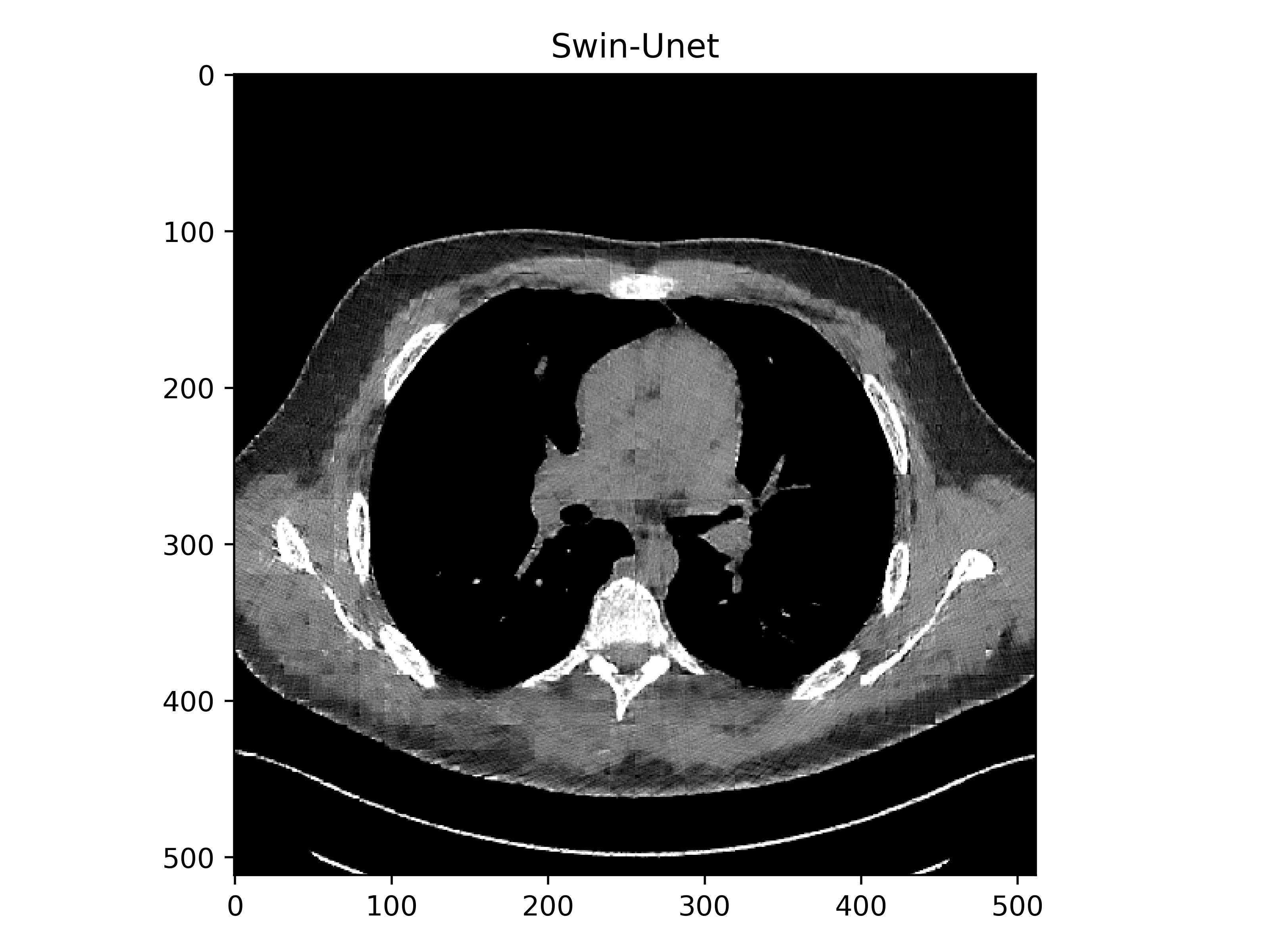}
	}
	\quad
	\subfigure[]{
		\includegraphics[width=0.3\textwidth,trim=50 10 80 10, clip]{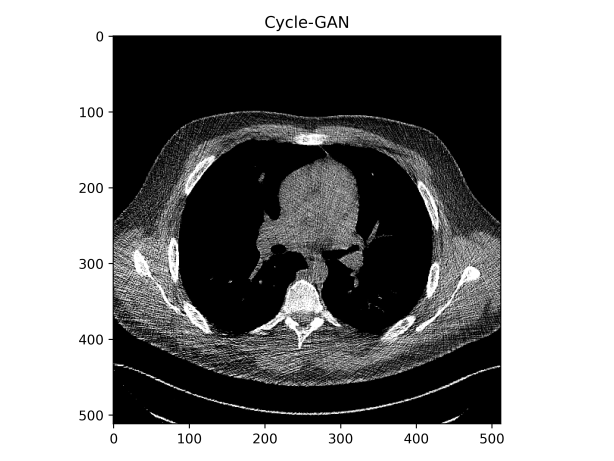}
	}
	\subfigure[]{
		\includegraphics[width=0.3\textwidth,trim=50 10 80 10, clip]{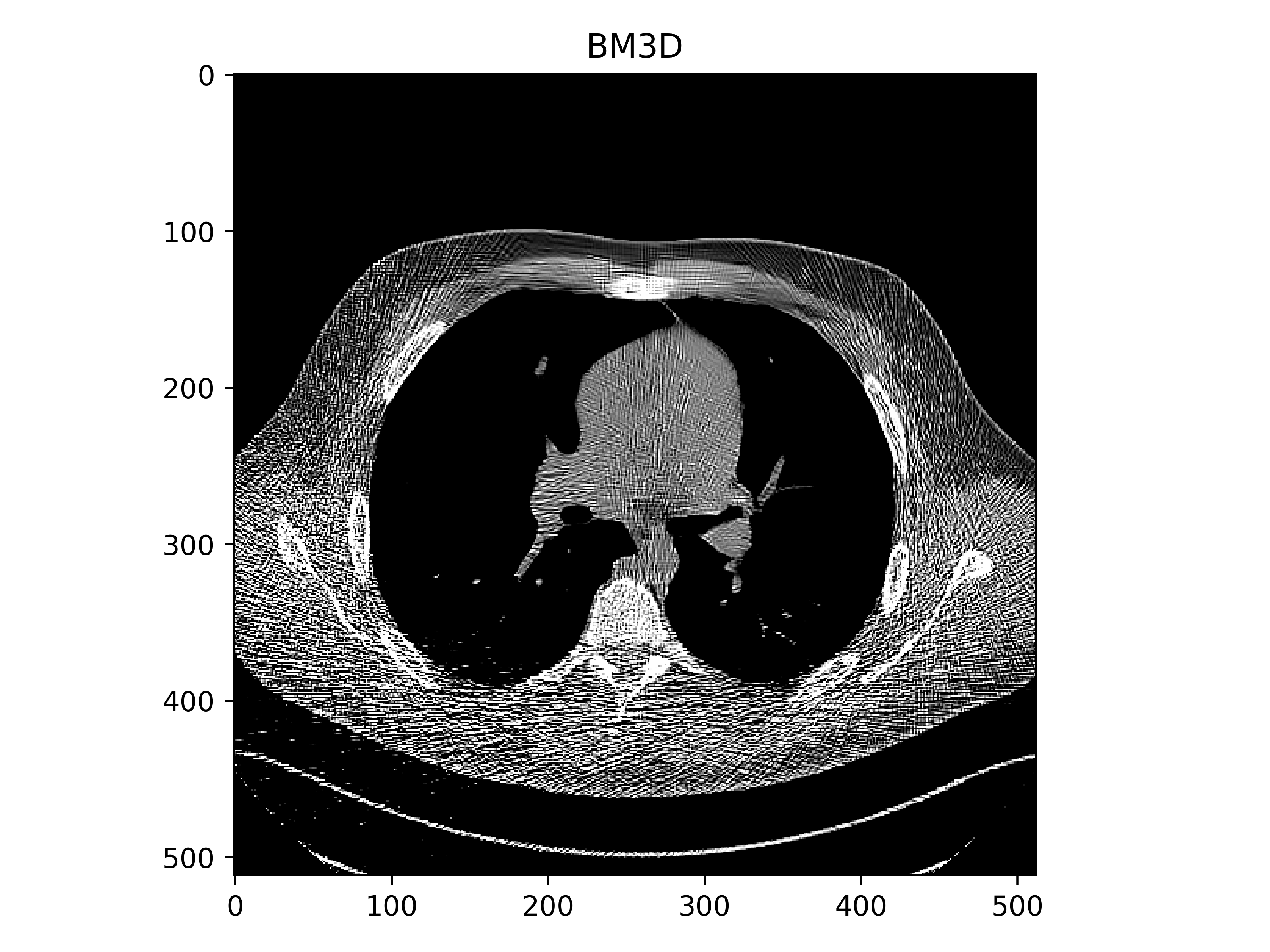}
	}
	\subfigure[]{
		\includegraphics[width=0.3\textwidth,trim=50 10 80 10, clip]{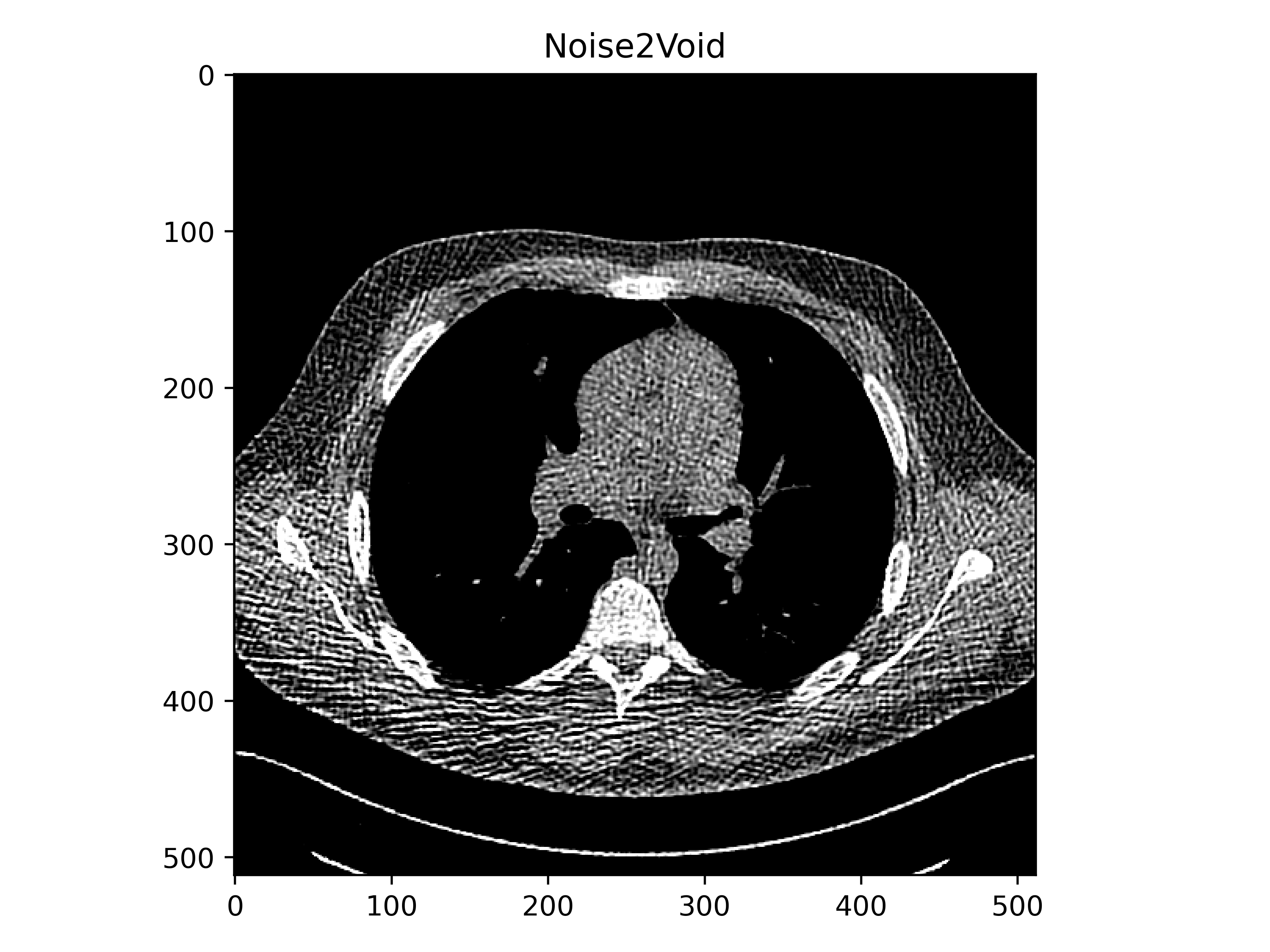}
	}
	\quad
	\subfigure[]{
		\includegraphics[width=0.3\textwidth,trim=50 10 80 10, clip]{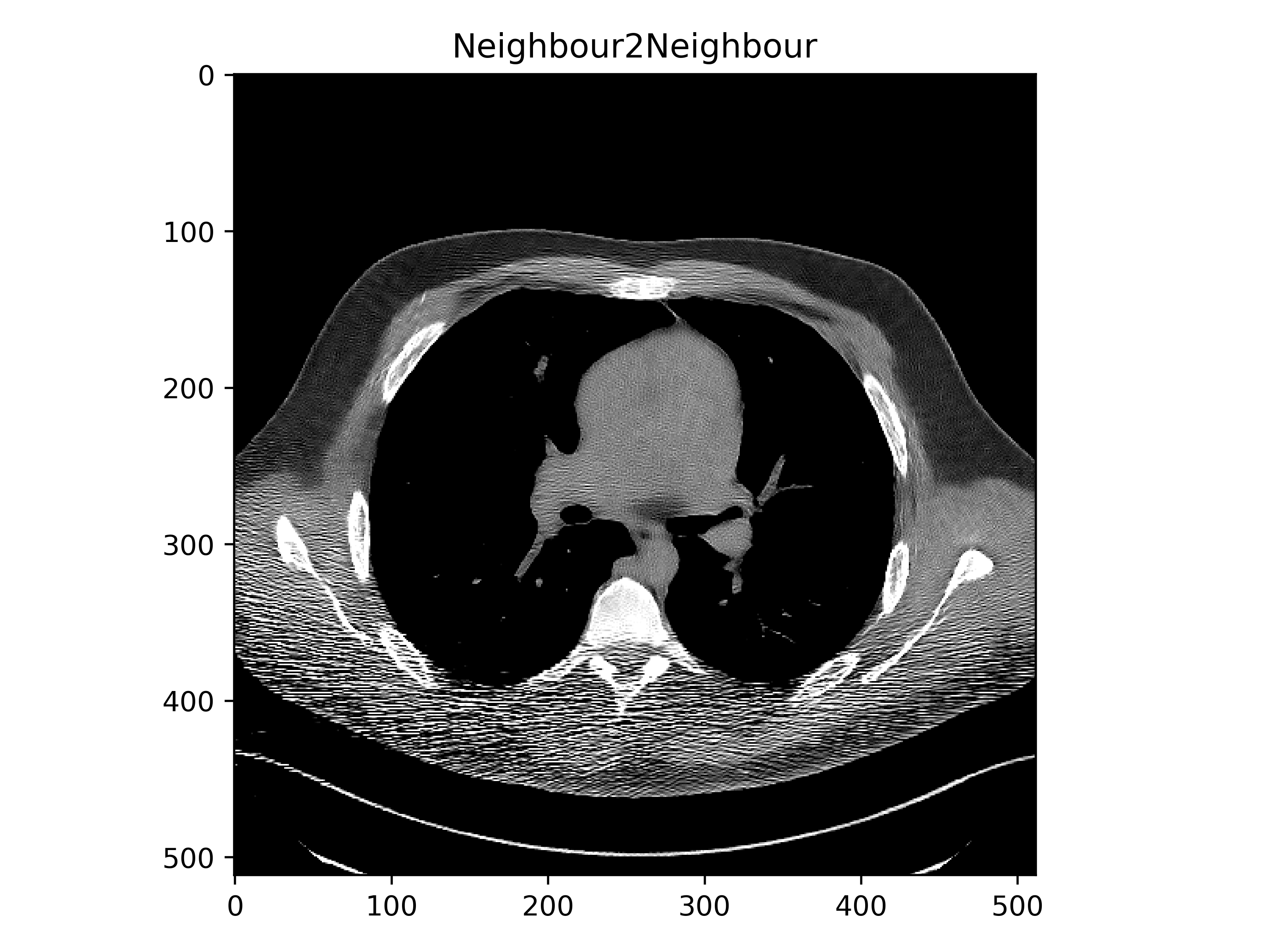}
	}
	\subfigure[]{
		\includegraphics[width=0.3\textwidth,trim=50 10 80 10, clip]{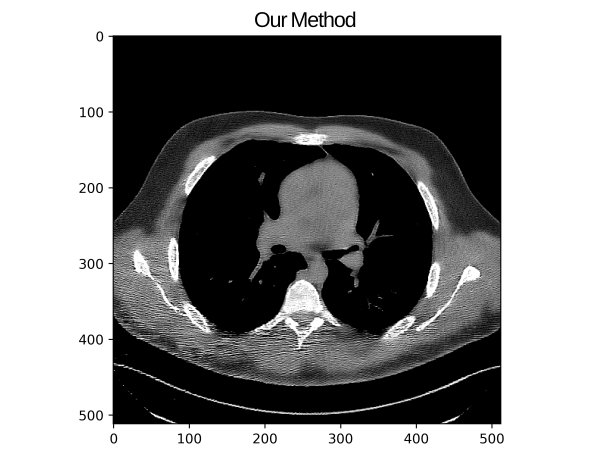}
	}
	\subfigure[]{
		\includegraphics[width=0.3\textwidth,trim=50 10 80 10, clip]{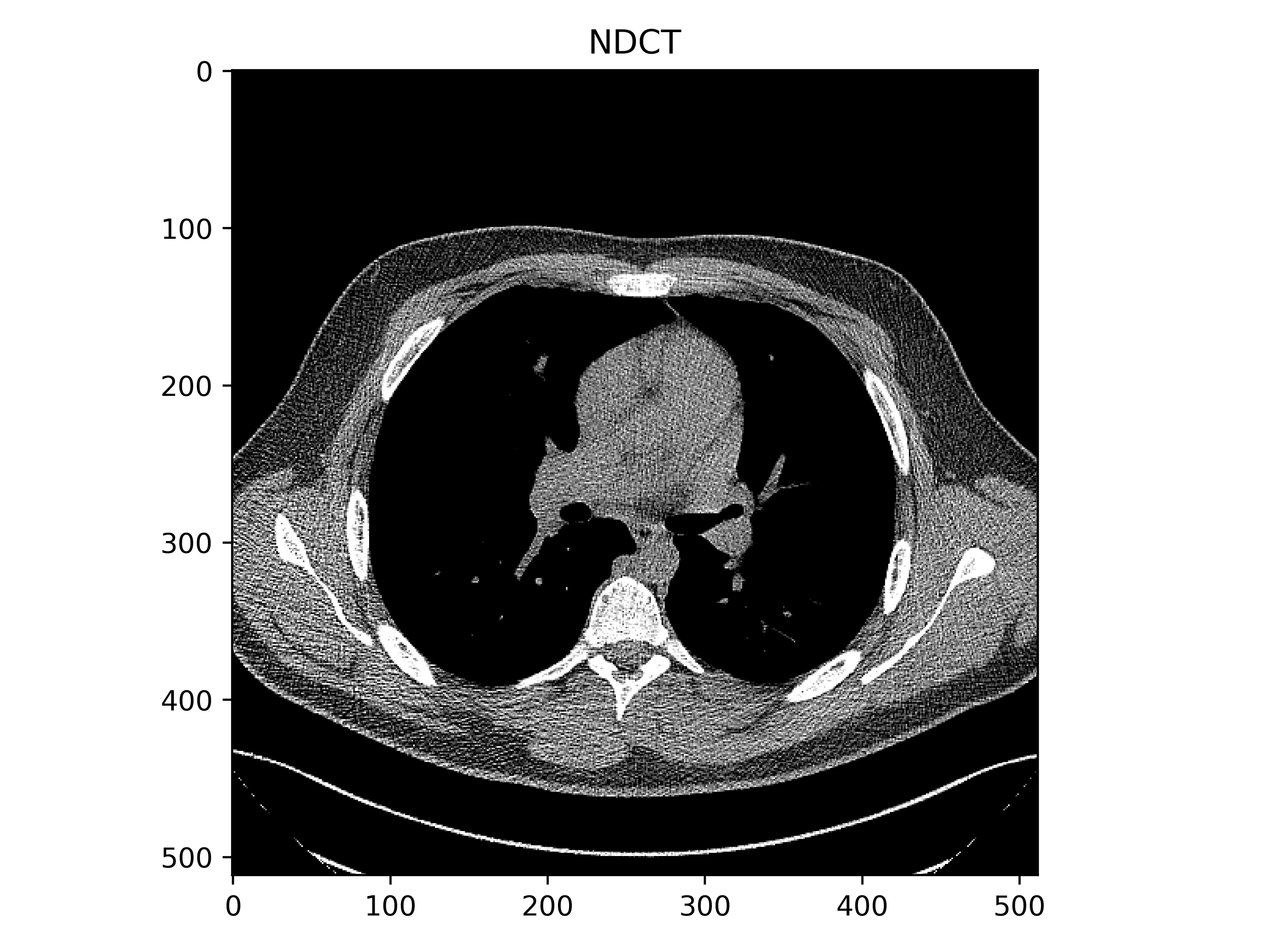}
	}
	
	\caption{Results of chest image for comparison w.r.t. NDCT. (a)LDCT$:$ 11.815dB, (b)RED-CNN$:$ 18.181dB, (c)Swin-Unet$:$ 17.804dB, (d)Cycle-GAN$:$ 15.073dB, (e)BM3D$:$ 12.923dB, (f)Noise2Void$:$ 14.095dB, (g)Neighbor2Neighbor$:$ 15.838dB, (h)PMRL$+$Unet$:$ 16.090dB, (I)Full-dose (FD)CT/NDCT.}
	\label{LDCT_chest_fig}
\end{figure}

To further demonstrate the effectiveness of the proposed method, we present representative pelvis and chest slices in Fig. \ref{LDCT_fig} and Fig. \ref{LDCT_chest_fig}, respectively. The proposed approach achieves superior noise suppression while preserving sharp anatomical boundaries and fine structural details, yielding results that are visually closest to the NDCT reference. In contrast, other methods, particularly self-supervised approaches, as well as certain supervised models such as Swin-Unet and Cycle-GAN, exhibit noticeable residual noise or introduce smoothing artifacts that degrade structural fidelity. It demonstrates that our progressive masked refinement learning effectively balances noise removal and detail preservation without relying on paired NDCT data.

\begin{table}[t]
\centering
\begin{tabular}{l c c cc}
\cline{1-5}
Methods& PSNR &SSIM & RMSE  & parameters\\
\cline{1-5}
LDCT (Dataset) & 12.272 & 0.579 & 100.705  &-\\
\cline{1-5}
\multicolumn{5}{c}{\textit{Supervised}}         \\ \hdashline
RED-CNN & \textbf{19.792} & \textbf{0.670} & \textbf{43.079}& 1.85M\\
Swin-Unet & 19.269&0.651& 45.497&  0.95M\\
Cycle GAN & 16.149& 0.643 &64.333 & 114M\\
\cline{1-5}
\multicolumn{5}{c}{\textit{Self-supervised}}         \\ \hdashline 
BM3D&14.476&0.626&82.430&-\\
Noise2Void& 15.332 &0.596& 75.088&31M\\
Neighbor2Neighbor&  17.312&\textbf{0.640}& 57.351&31M\\
PMRL+Unet &\textbf{18.131} & 0.636&\textbf{52.458} &31M\\
\cline{1-5}
\end{tabular}
\caption{Quantitative evaluation for LDCT chest dataset. PMRL denotes the proposed Progressive Masked Refinement Self-supervised learning.}\label{comparison_chest}
\end{table}

\subsubsection{Comparison Regarding Patients}
\begin{figure}[H]
\centering
\includegraphics[height=93px]{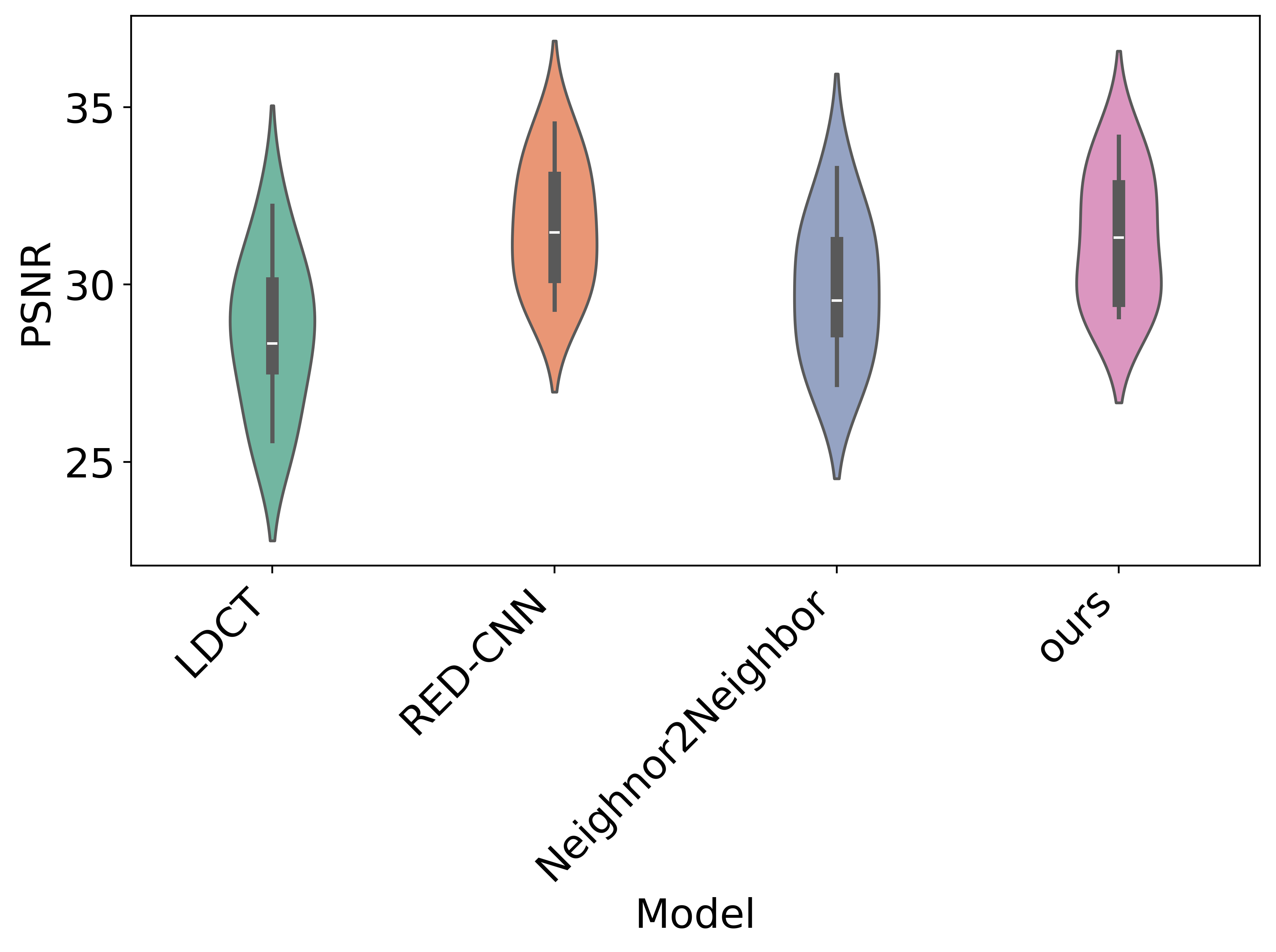}
\includegraphics[height=93px]{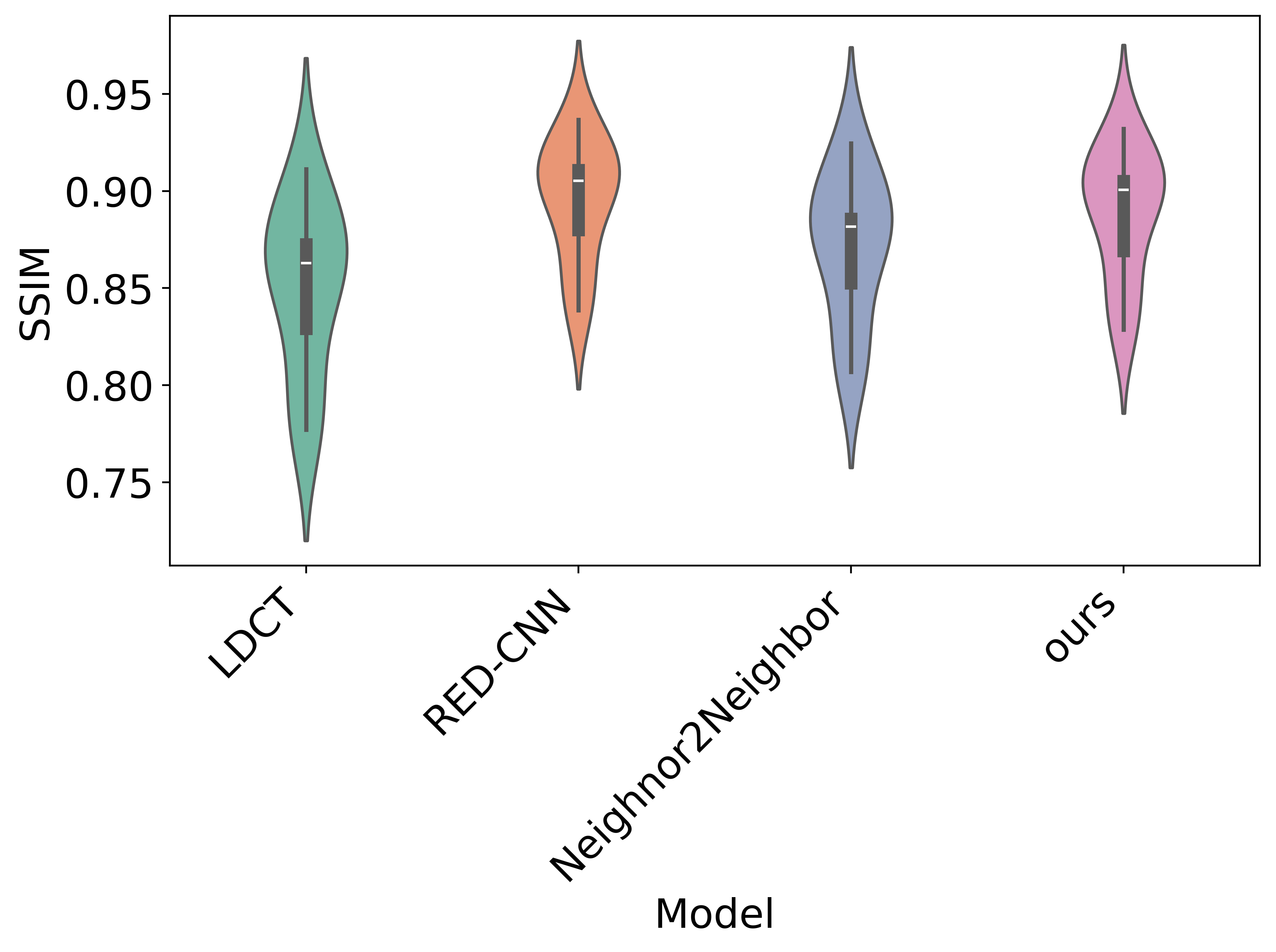}
\includegraphics[height=93px]{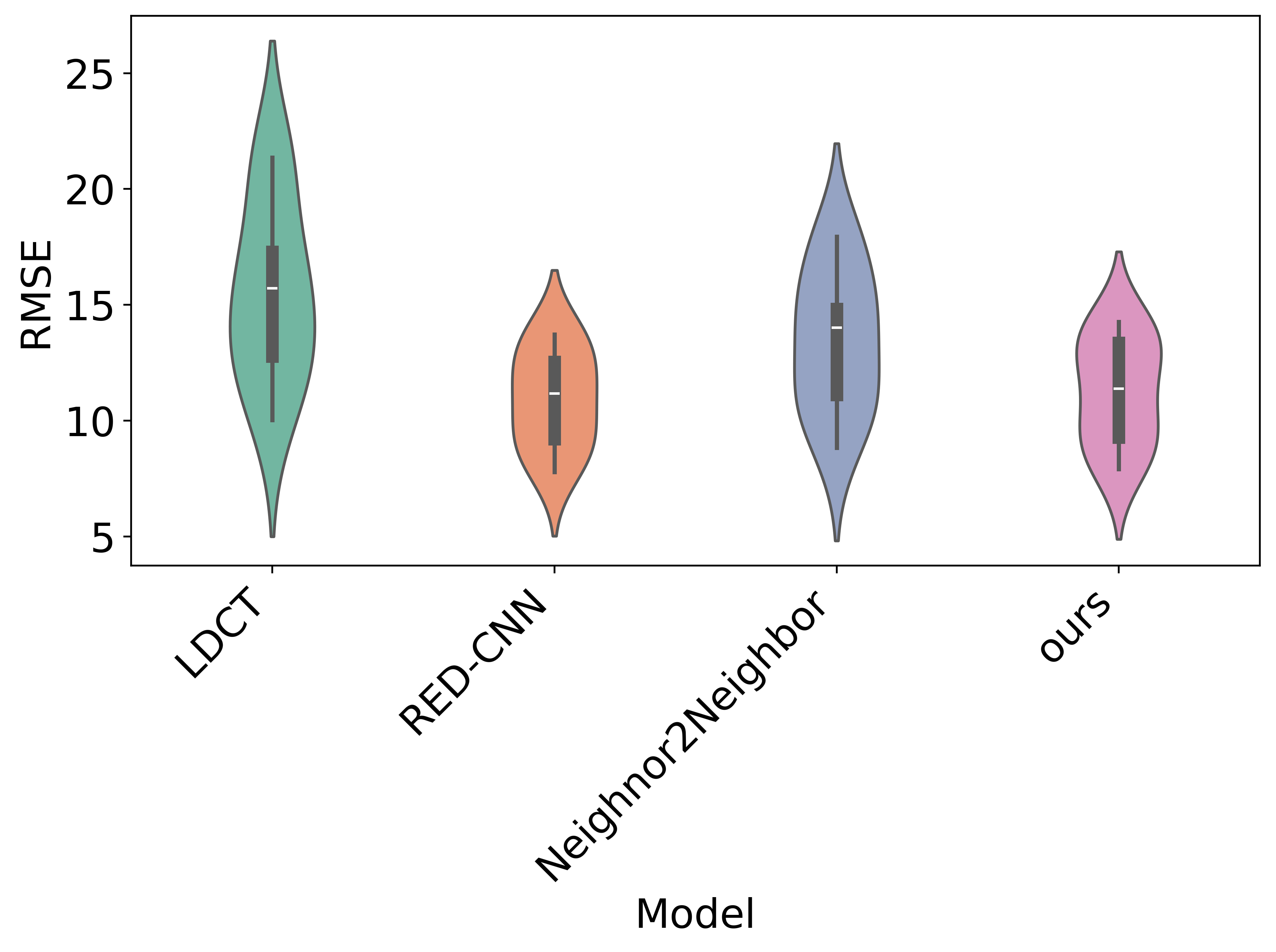}
\caption{Quantitative results on the abdomen test set of different patients. RED-CNN is the best-performing supervised method, while Neighbor2Neighbor is the strongest self-supervised baseline, as reported in Table \ref{comparison}.}\label{patient}
\end{figure}

We further assess the robustness of our method across 9 different patients in the evaluation split. Fig. \ref{patient} presents a quantitative comparison of LDCT, a representative supervised method (RED-CNN), a representative self-supervised method (Neighbor2Neighbor), and our approach in terms of PSNR, SSIM, and RMSE by averaging denoising results of each patient. The results show that our method exhibits substantially lower variance than LDCT and Neighbor2Neighbor, particularly in SSIM and RMSE, indicating more consistent reconstruction quality across patients. In addition, our method achieves performance comparable to supervised RED-CNN across all evaluation metrics. Notably, our approach outperforms LDCT for all patients, yielding average improvements of 2.663dB in PSNR, 0.038 in SSIM, and a 4.266 reduction in RMSE. These gains are maintained even in cases where the initial LDCT images already exhibit relatively high PSNR values.

\subsection{Ablation study}

\subsubsection{Combination of Noise}
We further analyze the effects of noise combination with the proposed progressive masked refinement learning. The results are summarized in Table \ref{ablation}. The ablation results demonstrate the complementary contributions of combination noise and progressive masked refinement learning. Following the previous study on mixed-noise for training \cite{zhu2025self}, introducing combination noise alone improves denoising performance by providing more effective training perturbations and reducing overfitting. Importantly, incorporating progressive masked refinement learning into this configuration further improves PSNR from 30.242 dB to 31.510 dB, demonstrating its additional contribution to the overall framework. This improvement suggests that progressive refinement enables the model to benefit further from the noise-injected training process. The full model achieves the best performance, supporting the combined use of both strategies.


\begin{table}[t]
\centering
\begin{tabular}{l c c c c}
\cline{1-4}
Metric & LDCT &add combination noise &  add progressive masked denoise\\
\cline{1-4}
PSNR& 28.842 &30.242& 31.510\\
SSIM& 0.856 &0.877& 0.892\\
RMSE& 15.160&12.718& 10.993\\
\cline{1-4}
\end{tabular}
\caption{Results of ablation study for both combination noise and progressive masked denoise strategy.}\label{ablation}
\end{table}

\subsubsection{Progressive masked refinement learning w.r.t. time step $k$}
\begin{figure}[H]
\centering
\includegraphics[width=0.8\textwidth]{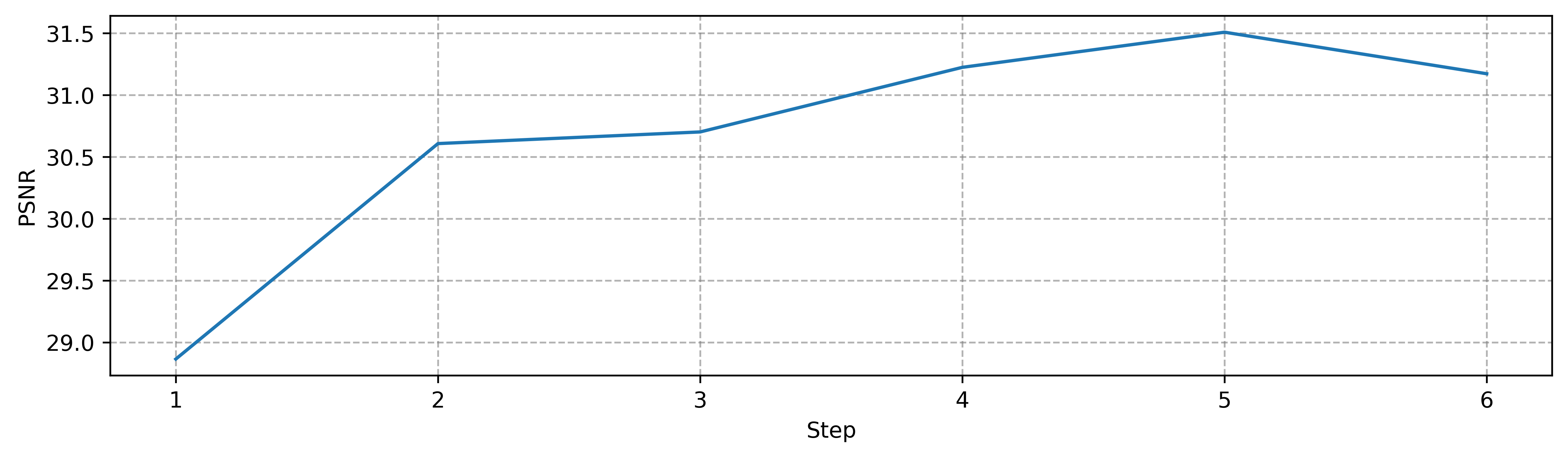}
\includegraphics[width=0.8\textwidth]{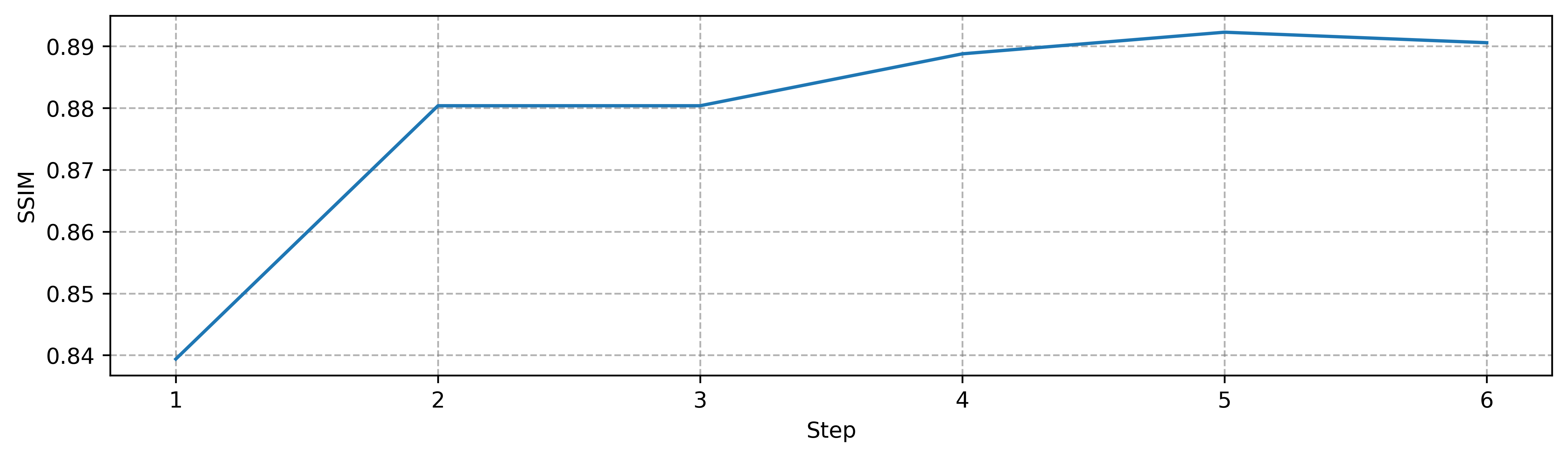}
\caption{Quantitative results on the validation set: number of time steps }\label{step}
\end{figure}

The number of time steps $k$ was selected empirically through a number of experiments. To ensure consistency between training and inference, the number of mask sampling steps during inference was also set to $k$. Note that the noise level and mask ratio were set to $\sigma=10$ (pixel intensity in the range of [0,255]) and $\alpha=0.1$, respectively. We evaluated $k$ in the range of 1 to 6. As shown in Fig. \ref{step}, the denoising performance consistently improves as $k$ increases and reaches its optimum at $k=5$. Further increasing $k$ leads to a degradation in performance in both PSNR and SSIM.
It indicates that from $k=1$ to $k=5$, the model learns to gradually denoise. Such behavior is expected, as more accurate recovery of corrupted regions requires sufficient contextual information from other parts of the image, which cannot be fully propagated in a single denoising step. By iteratively applying the denoising process, the denoised LDCT image at each step can be considered as augmented data for the next step. Thus, it maximizes the model's expressive power, leading to superior reconstruction without increasing the underlying parameter count. This assumption is consistent with the trend shown in Fig. \ref{step}, where performance improves with increased $k$ until sufficient contextual information has been accumulated, though the improvement becomes less pronounced between the 2-step and 3-step configurations.

\subsubsection{Noise level $\sigma$}
\begin{figure}[H]
\centering
\includegraphics[width=0.8\textwidth]{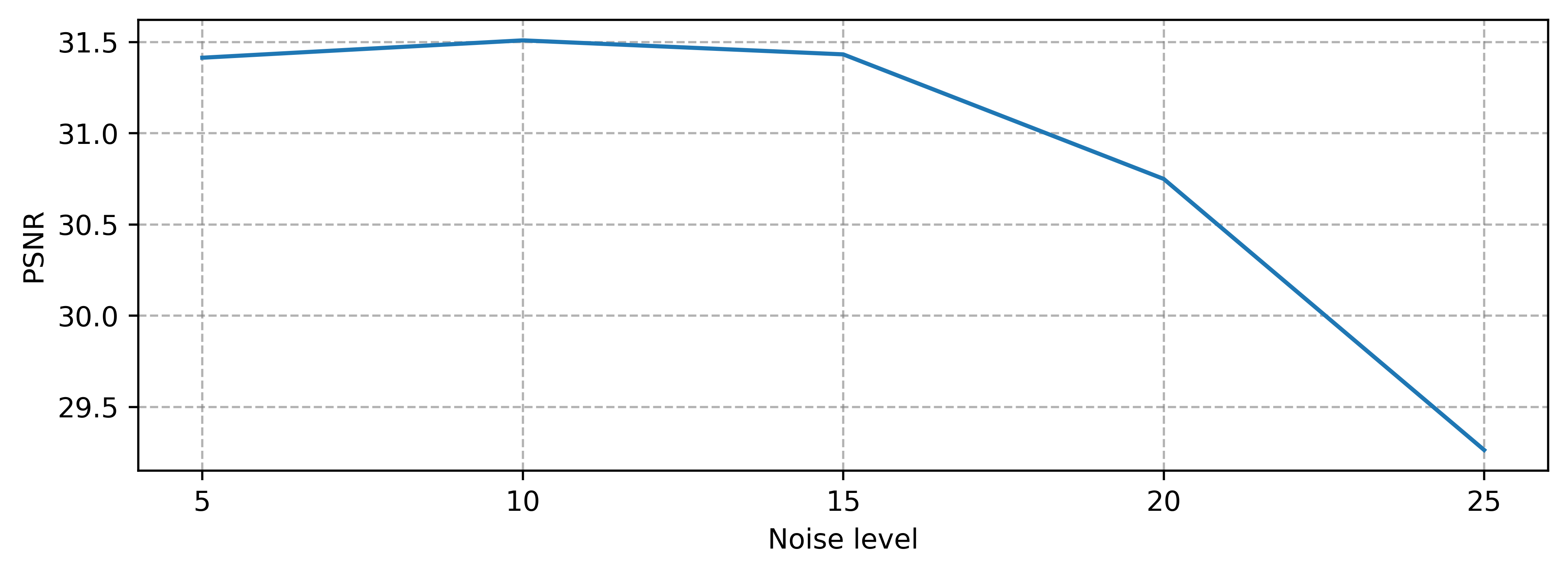}
\includegraphics[width=0.8\textwidth]{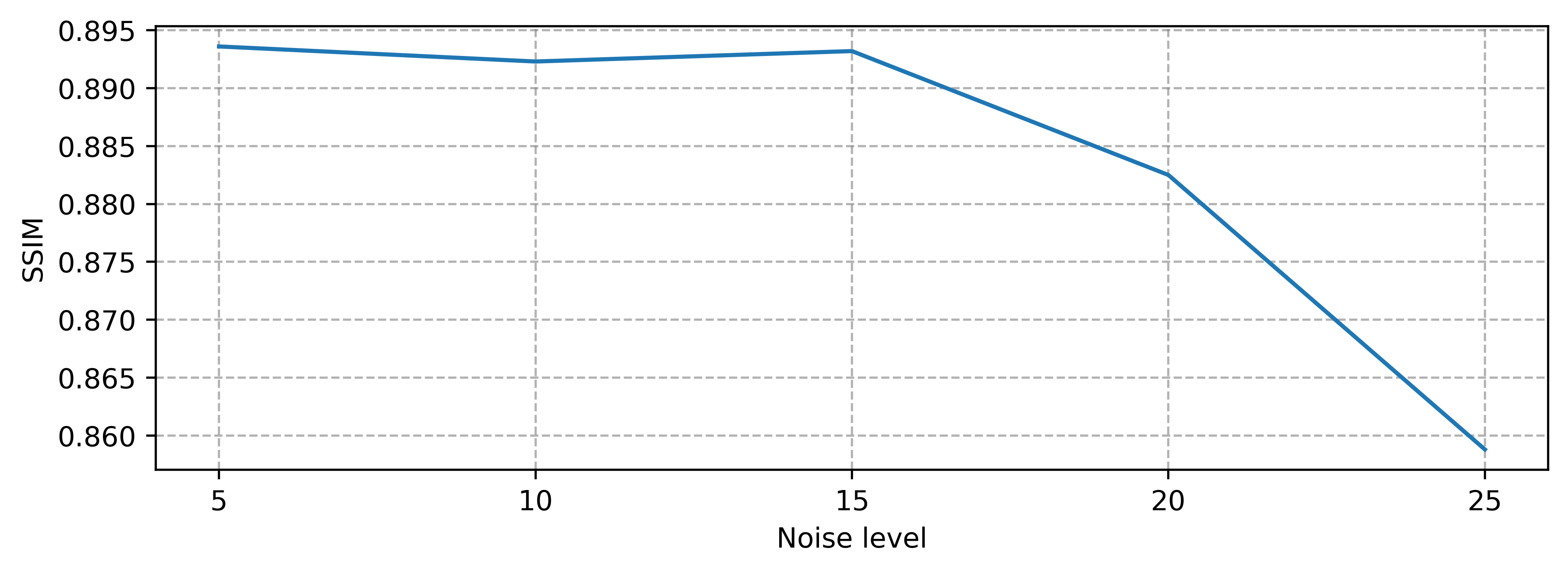}
\caption{Quantitative results on the validation set: noise level}\label{noise level}
\end{figure}

To investigate the impact of noise level on model performance, we evaluate different levels of the AWGN noise, where the noise level is controlled by the standard deviation of the Gaussian noise. We have added the noise before normalization and cropped the pixel to a maximum 255. Note that the time step and mask ratio were set to $k=5$ and $\alpha=0.1$, respectively. Fig. \ref{noise level} illustrates the effect of varying noise levels on performance over the test set. As the noise variance increases, both PSNR and SSIM consistently decrease, indicating degraded reconstruction quality. The highest PSNR is achieved at $\sigma=10$. For SSIM, the differences among noise levels $\sigma=5$, 10, and 15 are relatively minor, suggesting limited sensitivity within this range.

\subsubsection{Mask ratio $\alpha$}
\begin{figure}[H]
	\centering
	\includegraphics[width=0.8\textwidth]{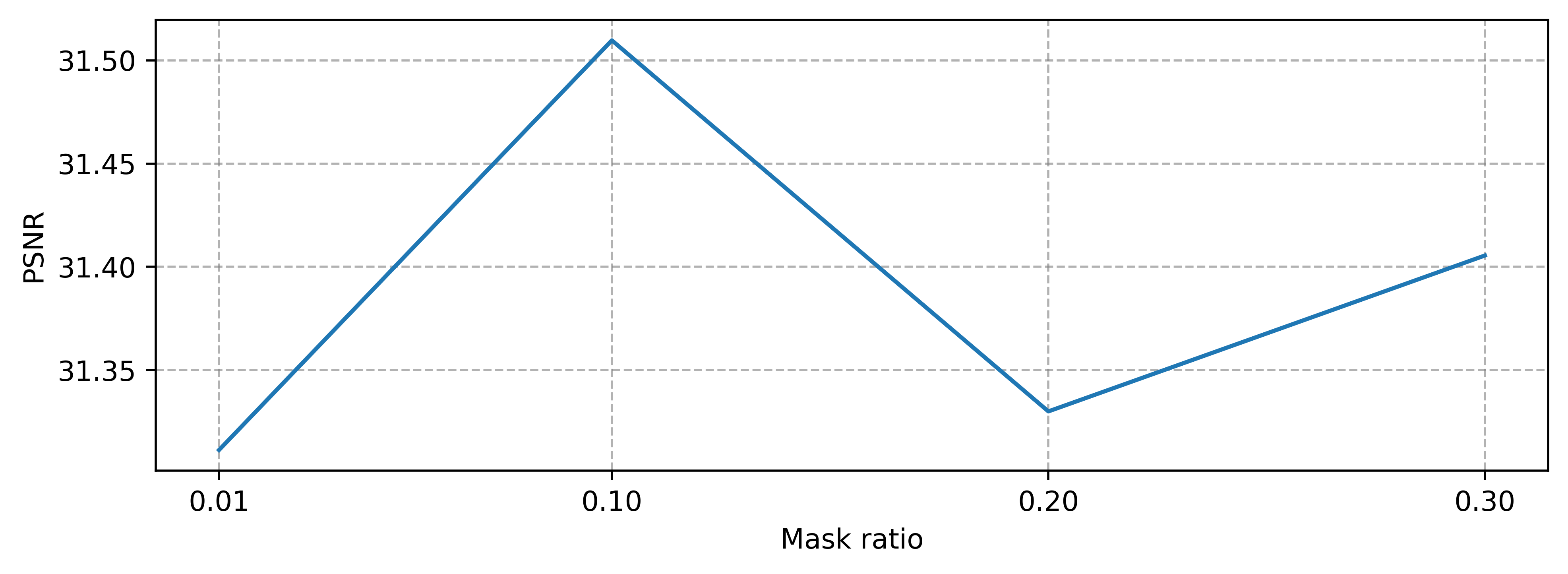}
	\includegraphics[width=0.8\textwidth]{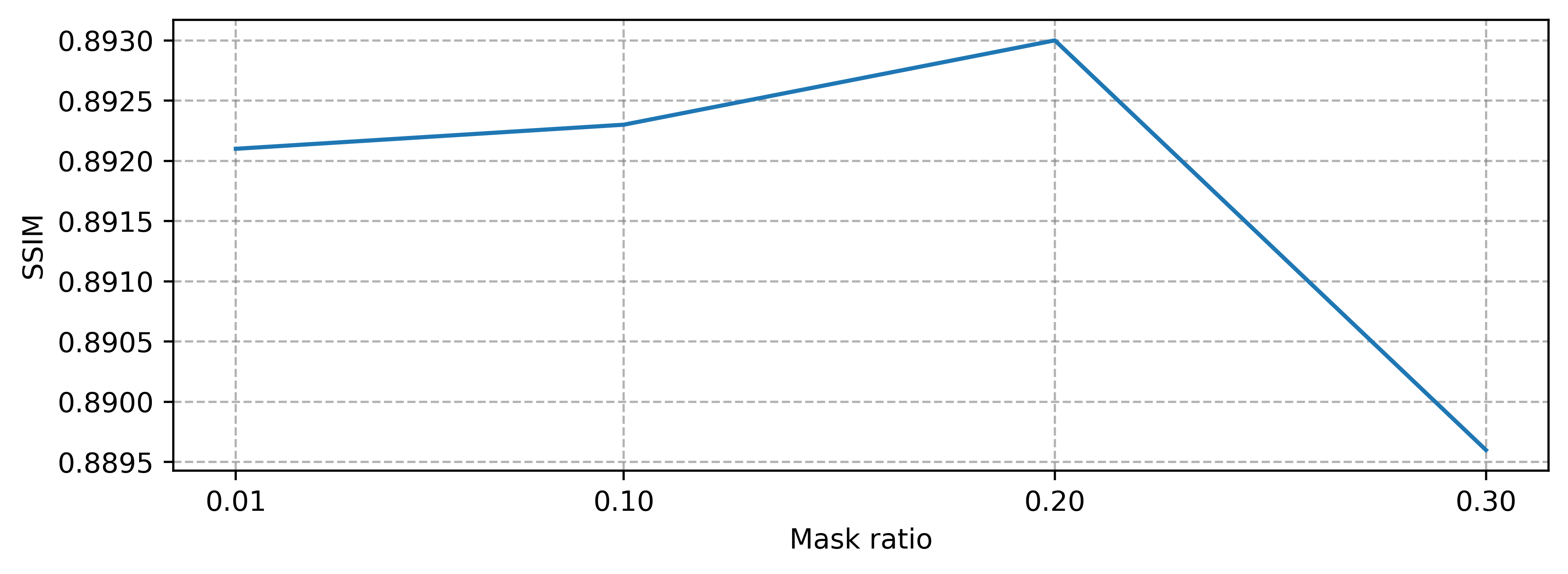}
	\caption{Quantitative results on the validation set: Mask ratio for progressive masked refinement learning}\label{mask ratio}
\end{figure}

We perform point-wise Bernoulli sampling over the spatial domain, where each pixel is independently activated with a success probability $1-\alpha$. Note that the noise level and time step were set to $\sigma=10$ and $k=5$, respectively. Fig. \ref{mask ratio} summarizes the impact of different masking ratios on denoising performance.
PSNR reaches its maximum at $\alpha=0.1$, while SSIM seems remains relatively stable over a broad range of masking ratios from $\alpha=0.01$ to $\alpha=0.2$. When the masking ratio is too low ($\alpha=0.01$), the supervision signal remains weak due to the high similarity between input and target, limiting performance improvements. In contrast, excessively high masking ratios ($\alpha=0.2$) removes substantial contextual information, resulting in degraded reconstruction quality and reduced structural fidelity.

\subsubsection{Consistency between training and inference strategies}
We also evaluated an inference strategy that follows the same progressive refinement procedure used during training. However, this approach resulted in substantially worse denoising performance than our proposed inference strategy, which applies the final trained model to multiple masked versions of the input and averages the resulting predictions. We attribute this difference to the distinction between training and inference: during training, the model parameters evolve through iterative optimization, whereas during inference, the parameters remain fixed at their final trained values. Therefore, applying the same progressive refinement procedure at inference time does not reproduce the model's evolution during training, which may explain the observed performance degradation.

\section{Discussion and Conclusions}

In this paper, we proposed progressive masked refinement learning framework that can improve the utilization of blind spot methods of individual LDCT images for denoising. This learning strategy incorporated three key elements: (a) unpaired, separate injection of combined noise for input and target, (b) random masking to reduce distribution discrepancy between synthetic training and original LDCT data, and (c) progressive training with different random maskes at each time step to reduce the under-utilization of blind spot methods. The incremental random masks applied on the inputs at increasing time steps introduced enhanced spatial correlation recovery. The controlled random noise in both the inputs and targets mitigated overfitting and strengthened the robustness of the self-supervised learning process. The evaluation on LDCT dataset, such as per-sample and patient-wise comparisons, demonstrated that the proposed method outperformed state-of-the-art self-supervised denoising approaches across multiple quantitative metrics. Moreover, our work exhibited consistent performance improvements across different backbone networks, suggesting its applicability to a range of architectures. These results indicated that progressive masked refinement learning significantly narrowed the performance gap between self-supervised and supervised denoising methods without increasing model complexity.

It is worth noting that Noise2Noise-based approaches typically require paired noisy observations with independent realizations of zero-mean noise, or assume specific additive noise distribution. These assumptions imply that LDCT images must follow specific noise statistics or share similar anatomical structures with paired training data, which limits their practical applicability. In contrast, Noise2Self and Noise2Void relax these assumptions and enable single-image self-supervised learning. However, their performance is often constrained by information loss due to sparse masking. Instead of relying on restrictive noise assumptions or paired data, our method extends the Noise2Self and Noise2Void paradigm by introducing a progressive masked refinement learning strategy. This allows each image sample to be more effectively utilized during training, progressively guiding the network output toward the NDCT target distribution. Another distinction from Noise2Void- and Noise2Self-based methods is that our approach does not rely on carefully engineered masking patterns or highly sensitive hyperparameter settings. For example, methods such as Noise2Void and Noise2Same require careful tuning of the learning rate (e.g., around $4\times10^{-4}$), whereas our method allows a standard training strategy. 


Despite these advantages, the proposed method has several limitations. In the ablation study, performance degrades when the number of time steps exceeds $k=5$. This is likely due to error accumulation across successive steps, which limits further improvement toward the NDCT target. Moreover, the evaluation is conducted exclusively on the LDCT Grand Challenge dataset, due to the limited availability of public LDCT datasets. Future work will extend this framework to other low-dose imaging modalities, such as low-dose positron emission tomography (LDPET), and investigate adaptive step-scheduling strategies. We also plan to assess its impact on downstream clinical tasks, including diagnosis and quantitative imaging.


\bibliographystyle{elsarticle-num}
\bibliography{ref}
\end{document}